%% file: main.tex
\documentclass[10pt,twocolumn,letterpaper]{article}

\usepackage{cvpr}              
\usepackage[accsupp]{axessibility}


\definecolor{cvprblue}{rgb}{0.21,0.49,0.74}
\usepackage[pagebackref,breaklinks,colorlinks,allcolors=cvprblue]{hyperref}

\def\paperID{*****} 
\def\confName{CVPR}
\def\confYear{2025}

\title{NTIRE 2025 challenge on Text to Image Generation Model Quality Assessment}

\author{Shuhao Han$^{*}$ \and Haotian Fan$^{*}$ \and Fangyuan Kong$^{*}$ \and Wenjie Liao$^{*}$ \and Chunle Guo$^{*}$ \and Chongyi Li$^{*}$ \and Radu Timofte$^{*}$ 
\and Liang Li$^{*}$ \and Tao Li$^{*}$ \and Junhui Cui$^{*}$ \and Yunqiu Wang$^{*}$ \and Yang Tai$^{*}$ \and Jingwei Sun\thanks{The organizers of the NTIRE 2025 challenge on Text to Image Generation Model Quality Assessment.\\
 The NTIRE 2025 website: \url{https://cvlai.net/ntire/2025/}.}
 \and Jianhui Sun \and Xinli Yue \and Tianyi Wang \and Huan Hou \and Junda Lu \and Xinyang Huang \and Zitang Zhou  \and Zijian Zhang \and Xuhui Zheng \and Xuecheng Wu \and Chong Peng \and Xuezhi Cao \and Trong-Hieu Nguyen-Mau \and Minh-Hoang Le \and Minh-Khoa Le-Phan \and Duy-Nam Ly \and Hai-Dang Nguyen \and Minh-Triet Tran \and Yukang Lin \and Yan Hong \and Chuanbiao Song \and Siyuan Li \and Jun Lan \and Zhichao Zhang \and Xinyue Li \and Wei Sun \and Zicheng Zhang \and Yunhao Li \and Xiaohong Liu \and Guangtao Zhai \and Zitong Xu \and Huiyu Duan \and Jiarui Wang \and Guangji Ma \and Liu Yang \and Lu Liu \and Qiang Hu \and Xiongkuo Min \and Zichuan Wang \and Zhenchen Tang \and  Bo Peng \and Jing Dong \and Fengbin Guan \and Zihao Yu \and Yiting Lu \and Wei Luo \and Xin Li \and Minhao Lin \and Haofeng Chen \and Xuanxuan He \and Kele Xu \and Qisheng Xu \and Zijian Gao \and Tianjiao Wan \and Bo-Cheng, Qiu \and Chih-Chung Hsu \and Chia-ming Lee \and Yu-Fan Lin \and Bo Yu \and Zehao Wang \and Da Mu \and Mingxiu Chen \and JunkangFang \and Huamei Sun \and Wending Zhao \and Zhiyu Wang \and Wang Liu \and Weikang Yu \and Puhong Duan \and Bin Sun \and Xudong Kang \and Shutao Li \and Shuai He \and Lingzhi Fu \and Heng Cong \and Rongyu Zhang \and Jiarong He \and Zhishan Qiao \and Yongqing Huang \and Zewen Chen \and  Zhe Pang \and Juan Wang \and Jian Guo \and Zhizhuo Shao \and Ziyu Feng \and Bing Li \and  Weiming Hu \and Hesong Li \and Dehua Liu \and Zeming Liu \and Qingsong Xie \and Ruichen Wang \and Zhihao Li \and Yuqi Liang \and Jianqi Bi \and Jun Luo \and Junfeng Yang \and Can Li \and Jing Fu \and Hongwei Xu \and Mingrui Long \and Lulin Tang
}

\begin{document}
\maketitle
\input{sec/0_abstract}    
\input{sec/1_intro}
\input{sec/2_related}

\input{sec/3_challenge}
\input{sec/4_result}
\input{sec/5_methods}

\section*{Acknowledgments}
This work was partially supported by the Humboldt Foundation. We thank the NTIRE 2025 sponsors: ByteDance, Meituan, Kuaishou, and University of Wurzburg (Computer Vision Lab).

\input{sec/X_appendix}
{
    \small
    \bibliographystyle{ieeenat_fullname}
    \bibliography{main}
}


\end{document}

%% file: sec/0_abstract.tex
\begin{abstract}
This paper reports on the NTIRE 2025 challenge on Text to Image (T2I) generation model quality assessment, which will be held in conjunction with the New Trends in Image Restoration and Enhancement Workshop (NTIRE) at CVPR 2025. 
The aim of this challenge is to address the fine-grained quality assessment of text-to-image generation models. 
This challenge evaluates text-to-image models from two aspects: image-text alignment and image structural distortion detection, and is divided into the alignment track and the structural track.
The alignment track uses the EvalMuse-40K, which contains around 40K AI-Generated Images (AIGIs) generated by 20 popular generative models.
The alignment track has a total of 371 registered participants. A total of 1,883 submissions are received in the development phase, and 507 submissions are received in the test phase. Finally, 12 participating teams submitted their models and fact sheets.
The structure track uses the EvalMuse-Structure, which contains 10,000 AI-Generated Images (AIGIs) with corresponding structural distortion mask. A total of 211 participants have registered in the structure track. A total of 1155 submissions are received in the development phase, and 487 submissions are received in the test phase. Finally, 8 participating teams submitted their models and fact sheets. Almost all methods have achieved better results than baseline methods, and the winning methods in both tracks have demonstrated superior prediction performance on T2I model quality assessment.
\end{abstract}

%% file: sec/1_intro.tex
\section{Introduction}
\label{sec:intro}
With the rapid development of generative models, advanced text-to-image (T2I) models are capable of generating many impressive images. 
However, these generated images still face challenges in terms of alignment with the text and structural fidelity.
Currently, widely used benchmarks~\cite{AIGIQA-20K,wang2023aigciqa2023,zhang2023perceptual,li2023agiqa} and methods~\cite{zhang2020blind,sun2023blind,zhang2023liqe,wu2023qalign} for evaluating the quality of generated images primarily focus on annotating and predicting image quality using Mean Opinion Scores (MOS).  
To provide a more comprehensive evaluation, certain benchmarks, such as AIGCIAQ2023~\cite{wang2023aigciqa2023}, assess images across multiple dimensions, including quality, authenticity and correspondence.
For images with low text alignment scores, it is often difficult to identify which elements of the text are not reflected in the generated images. 
Similarly, for images with low authenticity scores, pinpointing specific locations of structural distortion within the image remains challenging. 
Thus, developing more effective methods for assessing image-text alignment and detecting structural degradation in generated images is important.

This NTIRE 2025 Text to Image Generation Model Quality Assessment Challenge aims to promote the development of the methods for predicting quality scores of generated images while enabling fine-grained evaluation, thereby guiding the performance improvement of generative models.
This challenge is divided into the alignment track and structure track. 
For alignment track, we use EvalMuse-40K~\cite{han2024evalmuse}, which consists of around 40K image-text pairs with fine-grained alignment annotations.
For structure track, we construct a new dataset called EvalMuse-Structure, which includes 10K generated images.
Each image is provided with a structure score about image authenticity as well as a corresponding structural annotation mask that indicates whether structural distortions occur at specific locations within the image.

This is the first time that a fine-grained generated image quality assessment challenge has been held at NTIRE workshop.
The challenge has a total 582 registered participants, 371 in the alignment track and 211 in the structure track.
A total of 3038 submissions were received in the development phase, while 994 prediction results were submitted during the final test phase. 
Finally, 16 valid participating teams in the image track and 12 valid participating teams in the video track submitted their final models and fact sheets. 
They have provided detailed introductions to their IQA methods for fine-grained alignment and structural evaluation. 
We provide detailed results of the challenge in Section 4, and describe the specific methods used by the participating teams in Section 5.
We hope that this challenge can promote the development of fine-grained quality evaluation for generated images and guide improvements in T2I models, particularly in terms of image-text alignment and structural fidelity.


This challenge is one of the NTIRE 2025~\footnote{\url{https://www.cvlai.net/ntire/2025/}} Workshop associated challenges on: ambient lighting normalization~\cite{ntire2025ambient}, reflection removal in the wild~\cite{ntire2025reflection}, shadow removal~\cite{ntire2025shadow}, event-based image deblurring~\cite{ntire2025event}, image denoising~\cite{ntire2025denoising}, XGC quality assessment~\cite{ntire2025xgc}, UGC video enhancement~\cite{ntire2025ugc}, night photography rendering~\cite{ntire2025night}, image super-resolution (x4)~\cite{ntire2025srx4}, real-world face restoration~\cite{ntire2025face}, efficient super-resolution~\cite{ntire2025esr}, HR depth estimation~\cite{ntire2025hrdepth}, efficient burst HDR and restoration~\cite{ntire2025ebhdr}, cross-domain few-shot object detection~\cite{ntire2025cross}, short-form UGC video quality assessment and enhancement~\cite{ntire2025shortugc,ntire2025shortugc_data}, text to image generation model quality assessment~\cite{ntire2025text}, day and night raindrop removal for dual-focused images~\cite{ntire2025day}, video quality assessment for video conferencing~\cite{ntire2025vqe}, low light image enhancement~\cite{ntire2025lowlight}, light field super-resolution~\cite{ntire2025lightfield}, restore any image model (RAIM) in the wild~\cite{ntire2025raim}, raw restoration and super-resolution~\cite{ntire2025raw} and raw reconstruction from RGB on smartphones~\cite{ntire2025rawrgb}.

%% file: sec/2_related.tex
\section{Related Work}
\label{sec:related}

\subsection{AIGI dataset}
In recent years, several AI-Generated Images (AIGI) datasets have been proposed. 
Benefiting from the success of Stable Diffusion~\cite{he2022latent}, DiffusionDB~\cite{wang2022diffusiondb} has collected 14 million images generated by Stable Diffusion based on prompts and hyperparameters provided by real users. 
HPS~\cite{wu2023human} and Pick-A-Pic~\cite{kirstain2024pick} collect a large number of side-by-side image comparisons to evaluate the quality of the generated images. 
AGIQA-1K~\cite{zhang2023perceptual}, AGIQA-3K~\cite{li2023agiqa}, AIGCIQA2023~\cite{wang2023aigciqa2023} and AIGIQA-20K~\cite{AIGIQA-20K} annotate the quality of the generated images with scores and get MOSs to assess the quality of the generated images. 
GenAI-Bench~\cite{li2024genai} provides score-based annotations to evaluate the alignment between generated images and text for overall image-text alignment.
Gecko~\cite{wiles2024revisiting} and RichHF~\cite{liang2024rich} enable fine-grained evaluation by annotating the inconsistent words in the text relative to the generated images. 
Additionally, RichHF employs point annotations on the generated images to highlight regions of distortion.
In this challenge, we utilize EvalMuse-40K~\cite{han2024evalmuse} for image-text alignment evaluation, which splits the text in each image-text pair and performs element-level annotations. 
Meanwhile, we use EvalMuse-Structure to assess the structural fidelity of generated images, which annotates structural distortions in the images using bounding boxes.

\subsection{IQA method}
Traditional IQA methods~\cite{zhang2020blind,sun2023blind, zhang2023liqe} focus on various image distortions such as noise, blur and semantic content. 
For generated images, consistency with the used text is considered as an important part of generated images quality evaluation.
HPS~\cite{wu2023human} and PickScore~\cite{kirstain2024pick} leverage the CLIP~\cite{radford2021clip} model to simulate human preferences for generated images.
ImageReward~\cite{xu2024imagereward} and FGA-BLIP2~\cite{han2024evalmuse} employ BLIP-based~\cite{li2022blip,li2023blip2} architectures to predict scores.
"With the advancement of Multi-modal Large Language Models (MLLMs), VQAScore~\cite{lin2024evaluating} and Q-Align~\cite{wu2023qalign} evaluate image-text alignment and the quality of generated images by employing visual question answering on MLLMs.
For fine-grained image-text alignment evaluation, TIFA~\cite{hu2023tifa} and Gecko~\cite{wiles2024revisiting} generate specific questions targeting different elements in the text to evaluate the fine-grained alignment capabilities of the generated images. 
RAHF model~\cite{liang2024rich} performs fine-grained alignment evaluation by identifying inconsistent words and predicts heatmaps to assess structural distortions.

%% file: sec/3_challenge.tex
\section{NTIRE 2025 challenge on Text to Image Generation Model Quality Assessment}

This NTIRE 2025 Challenge on Text to Image Generation Model Quality Assessment is organized to improve the fined-grained quality assessment of generated images.
The main goal of this challenge is to achieve fine-grained alignment evaluation by predicting the alignment scores of elements and fine-grained structural evaluation by predicting structural distortion masks.
To accomplish these goals, the challenge is divided into alignment track and structure track. 
The details of the whole challenge are described in the following parts, including datasets, evaluation protocol and challenge phases.

\subsection{Datasets}
In the alignment track, we use the EvalMuse-40K~\cite{han2024evalmuse} dataset for training and validation. 
This dataset consists of around 40K image-text pairs with fine-grained alignment annotations, including 2K real prompts and 2K synthetic prompts. 
Real prompts are selected using an Mixed Integer Linear Programming (MILP) sampling strategy to ensure diversity and category balance. 
Synthetic prompts are generated by GPT-4 based on predefined templates targeting specific aspects such as quantity and spatial relationships. 
Furthermore, prompts are split into elements to achieve fine-grained annotations. During annotation, annotators provide an overall alignment score for each image-text pair and verify whether the split elements are represented in the generated images. 
Each pair is annotated by three annotators, and pairs with significant disagreements in alignment scores are re-annotated. 
During the testing phase, additional image-text pairs are introduced and annotated using the same process, which are then combined with a portion of the validation set as test dataset.

In the structure track, we create a new dataset, EvalMuse-Structure, for training, validation, and testing.
This dataset includes 12K generated images with fine-grained structural annotations, of which 10K are used for training, 1K for validation, and 1K for testing. 
For each generated image, annotators provide structural scores and use bounding boxes to annotate regions with structural distortions. 
Each image is annotated by three annotators, and for fine-grained annotations, structural distortion masks are derived from the overlapping distorted regions identified by at least two annotators.

\begin{table*}[t]
\centering
\caption{Quantitative results for NTIRE 2025 challenge on Text to Image Generation Model Quality Assessment: Track 1 Alignment.}
\label{tab:track1}
\begin{tabular}{c|c|c|cccc}
\toprule
Rank  & Team  & Leader & Main Score & SRCC  & PLCC  & ACC \\
\midrule
1     & IH-VQA & Jianhui Sun & 0.8551  & 0.8249  & 0.8485  & 0.8734  \\
2     & Evalthon & Zijian Zhang & 0.8426  & 0.8002  & 0.8321  & 0.8691  \\
3     & HCMUS & Trong-Hieu Nguyen-Mau & 0.8381  & 0.8101  & 0.8306  & 0.8559  \\
4     & MICV  & Jun Lan & 0.8221  & 0.7864  & 0.8050  & 0.8485  \\
5     & SJTU-MMLab & Zhichao Zhang & 0.8158  & 0.7729  & 0.8029  & 0.8438  \\
6     & SJTUMM & Zitong Xu & 0.8062  & 0.7563  & 0.7993  & 0.8346  \\
7     & WT    & Zichuan Wang & 0.7913  & 0.7413  & 0.7740  & 0.8249  \\
8     & YAG   & Fengbin Guan & 0.7777  & 0.7143  & 0.7456  & 0.8255  \\
9     & SPRank & Minhao Lin & 0.7604  & 0.6899  & 0.7280  & 0.8119  \\
10    & AIIG  & Xuanxuan He & 0.7386  & 0.6574  & 0.7073  & 0.7949  \\
11    & Joe1007 & Bo-Cheng, Qiu & 0.7359  & 0.6572  & 0.7041  & 0.7912  \\
12    & iCOST & Bo Yu & 0.7350  & 0.6630  & 0.7040  & 0.7865  \\
\midrule
Baseline & \multicolumn{2}{c|}{FGA-BLIP2~\cite{han2024evalmuse}} & 0.7256  & 0.6491  & 0.6947  & 0.7792  \\
\bottomrule
\end{tabular}%
\end{table*}

\begin{table*}
    \centering
    \caption{Quantitative results for NTIRE 2025 challenge on Text to Image Generation Model Quality Assessment: Track 2 Structure.}
\label{tab:track2}
\begin{tabular}{c|cc|cccc}
\toprule
Rank  & \multicolumn{1}{c|}{Team} & Leader & Main Score & F1 Score & SRCC  & PLCC \\
\midrule
1     & \multicolumn{1}{c|}{HNU-VPAI} & Zhiyu Wang & 0.6927  & 0.6938  & 0.6716  & 0.7085  \\
2     & \multicolumn{1}{c|}{OPDAI} & Shuai He & 0.6839  & 0.6709  & 0.6911  & 0.7374  \\
3     & \multicolumn{1}{c|}{MICV} & Jun Lan & 0.6832  & 0.6764  & 0.6735  & 0.7250  \\
4     & \multicolumn{1}{c|}{out of memory} & Qiao Zhishan & 0.6796  & 0.6652  & 0.6916  & 0.7352  \\
5     & \multicolumn{1}{c|}{I$^2$ Group} & Zewen Chen & 0.6716  & 0.6601  & 0.6731  & 0.7239  \\
6     & \multicolumn{1}{c|}{Brute Force Wins} & Zeming Liu & 0.6674  & 0.6590  & 0.6644  & 0.7095  \\
7     & \multicolumn{1}{c|}{Wecan EvalAIG} & Jun Luo & 0.6649  & 0.6535  & 0.6661  & 0.7165  \\
8     & \multicolumn{1}{c|}{Tenryu Babu} & Junfeng Yang      & 0.6161  & 0.5944  & 0.6423  & 0.6910  \\
\midrule
Baseline & \multicolumn{2}{c|}{RAHF model~\cite{liang2024rich}} & 0.6065 & 0.6021 & 0.5893 & 0.6444 \\
\bottomrule
\end{tabular}%
\end{table*}

\subsection{Evaluation protocol}
In both tracks, the main scores are utilized to determine the rankings of participating teams.
We evaluate the alignment scores and structure scores predicted by the model in both tracks using Spearman Rank-order Correlation Coefficient (SRCC) and Person Linear Correlation Coefficient (PLCC).
SRCC measures the prediction monotonicity, while PLCC measures the prediction accuracy. 
Better IQA methods should have larger SRCC and PLCC values. 
Before calculating PLCC index, we perform the third-order polynomial nonlinear regression.

In the alignment track, we use the accuracy (ACC) of the model in determining whether elements in the prompt are presented in the generated image to measure the model's fine-grained alignment evaluation capability.
The primary score computation method for the alignment track is as follows:
\begin{equation}
\mathrm{Main\;Score} = 0.5 * \mathrm{ACC} + 0.25 * ( \mathrm{SRCC} + \mathrm{PLCC})
\end{equation}

In the structure track, we evaluate the model's fine-grained structural assessment capability using the F1 Score computed between the model-predicted structural distortion mask and the human-annotated structural distortion mask. 
The primary score computation method for the structure track is as follows:
\begin{equation}
    \mathrm{Main\;Score} =0.7 *\mathrm{F1} + 0.15 * (\mathrm{SRCC} + \mathrm{PLCC})
\end{equation}

\subsection{challenge phases}
Both tracks consist of two phases: the development phase and the testing phase. 

\noindent \textbf{Development phase}: In this phase, participants have access to the training dataset, which contains generated images, corresponding prompts, and relevant annotations. 
Participants can learn about the structure of the dataset and develop their own methods. 
We also release a validation dataset that includes generated images and their corresponding prompts but lacks annotations.
Participants can use their methods to predict annotations for the validation dataset and upload the results to the server. 
Immediate feedback is provided, enabling participants to analyze the effectiveness of their methods on the validation set. 
The validation leaderboard is available.
\begin{figure*}[!t]
	\centering
	\includegraphics[width=\linewidth]{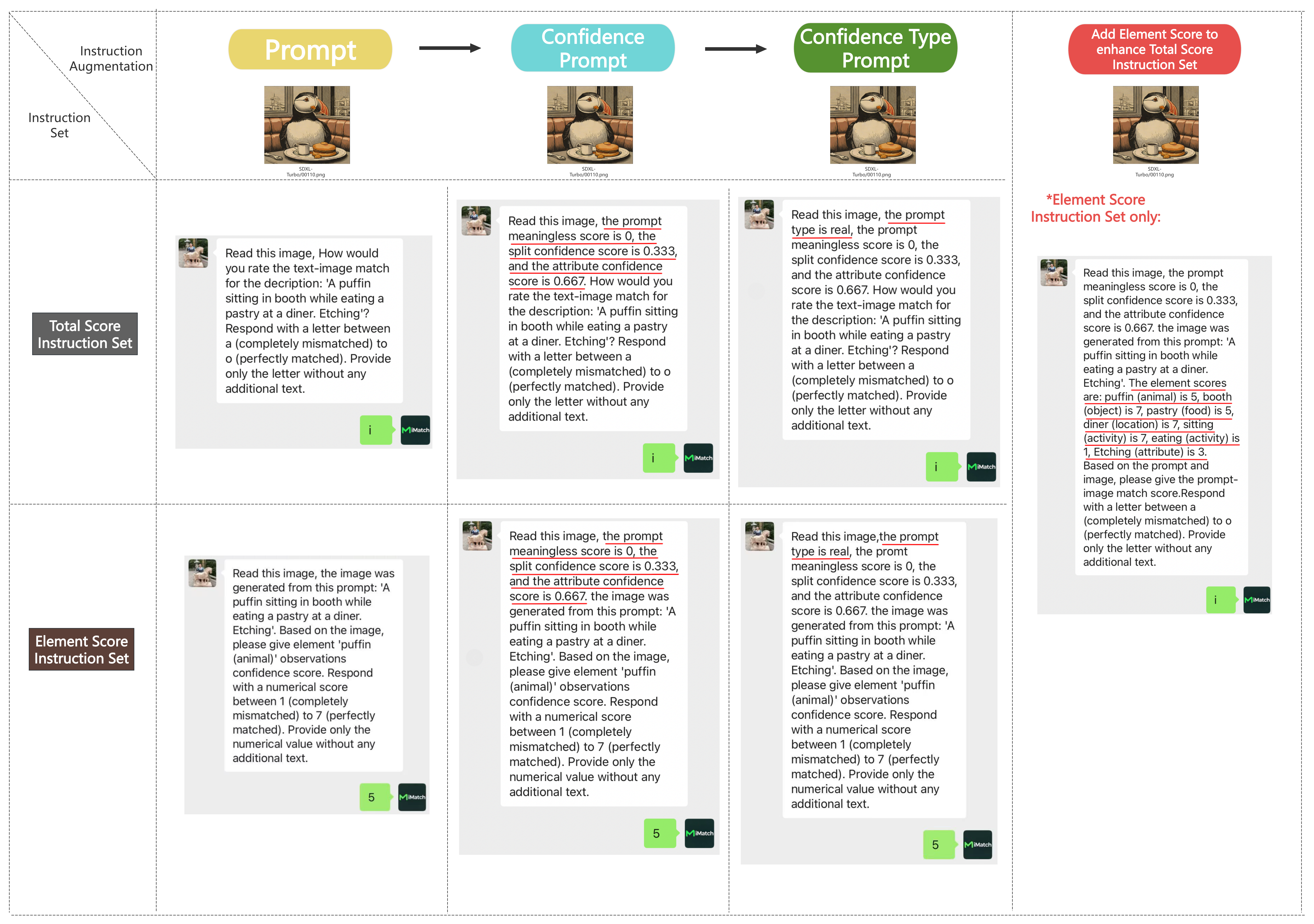}
	\caption{Overview of team IH-VQA proposed iMatch.}
	\label{fig_imatch}
\end{figure*}

\begin{figure}[!t]
    \centering
    \includegraphics[width=\linewidth]{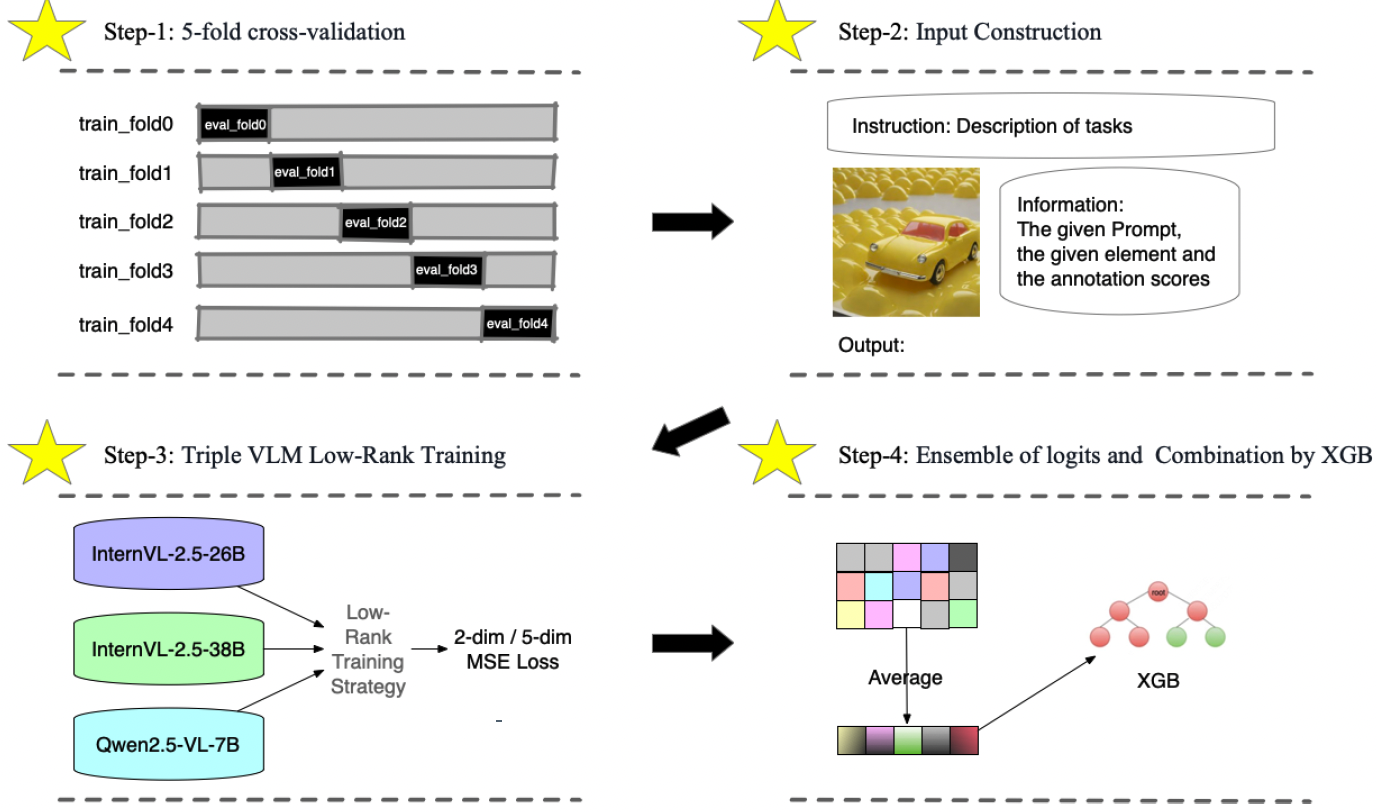}
    \caption{Overview of team Evalthon proposed method.}
    \label{fig:evalthon}
\end{figure}

\noindent \textbf{Testing phase:} 
In this phase, participants have access to the test dataset that contains generated images with the corresponding prompts, but without annotations. 
Participants need to upload the final predicted annotations of the test set before the challenge deadline. 
The leaderboard is not available at this stage and participants can only see their own scores.
At the same time, each team is required to submit source code/executable files and a fact sheet containing a detailed description of the proposed methodology and corresponding team information.
The final results and rankings are then sent to the participants.

%% file: sec/4_result.tex
\section{Challenge Results}
In this challenge, 12 teams in the alignment track and 8 teams in the structure track have submitted their final code/executables and fact sheets. Tab.~\ref{tab:track1} and~\ref{tab:track2} summarize the key results and important information for the 20 valid teams. These methods are briefly described in Section~\ref{sec:method} and the team members are listed in Appendix~\ref{sec_apd:track1_team} and~\ref{sec_apd:track2_team}.

In the alignment track, FGA-BLIP2 is used as the baseline, while RAHF model serves as the baseline for the structure track. 
Tab.~\ref{tab:track1} and ~\ref{tab:track2} reveal that all submitted results from participating teams achieve better performance than the baseline. 
In the alignment track, the top-performing team, IH-VQA, achieves 0.1295 higher Main Score than baseline, while six teams have Main Scores higher than 0.8. 
In the structure track, the leading team, HNU-VPAI, exceeds the baseline by 0.0862 in the Main Score. 
These methods achieve better fine-grained evaluations in both image-text alignment and structure distortion detection, which significantly advancing the development of fine-grained evaluation of generated images and facilitate better improvement of T2I generative models.

%% file: sec/5_methods.tex
\section{Challenge Methods}
\label{sec:method}
\subsection{Image-Text Alignment Track}

\subsubsection{IH-VQA}
\label{ih-vqa}

Team IH-VQA~\cite{yue2025instruction} wins the championship in the image-text alignment track with their proposed method, iMatch (Instruction-augmented Multimodal Alignment for Image-Text and Element Matching), as illustrated in \Cref{fig_imatch}. This method aims to precisely assess the alignment between generated images and textual descriptions through several key innovations. They fine-tune MLLMs using fine-grained image-text matching annotations from the EvalMuse-40K dataset to guide the model in learning nuanced correspondences. To further enhance performance, they propose four augmentation strategies: (1) the QAlign strategy, which maps textual rating levels to numerical scores and applies a soft mapping of prediction probabilities for more accurate score conversion; (2) the validation set augmentation strategy, where the model generates high-quality pseudo-labels on the validation set during training, which are then merged back into the training set to improve generalization; (3) the element augmentation strategy, which incorporates element labels into the user query, enabling Chain-of-Thought-style reasoning to derive better overall matching scores; and (4) the image augmentation strategy, introducing three augmentation techniques to increase the diversity of training images and enhance robustness to visual variations.

In the element matching task, they further propose two augmentation techniques: (1) prompt type augmentation, which embeds the prompt type (real or synthetic) into the query to help the model distinguish different source characteristics; and (2) score perturbation augmentation, which adds slight random noise to target labels to prevent overfitting and improve the model’s generalization.

Finally, they adopt a model ensemble strategy. For the image-text matching task, they average results from several fine-tuned MLLMs: Qwen2.5-VL-7B-Instruct\cite{qwen2.5vl}, Ovis2-8B, Ovis2-16B, and Ovis2-34B\cite{ovis}. For the element matching task, the ensemble combines Ovis2-8B and Ovis2-34B.
\begin{figure*}[!t]
    \centering
    \includegraphics[width=\linewidth]{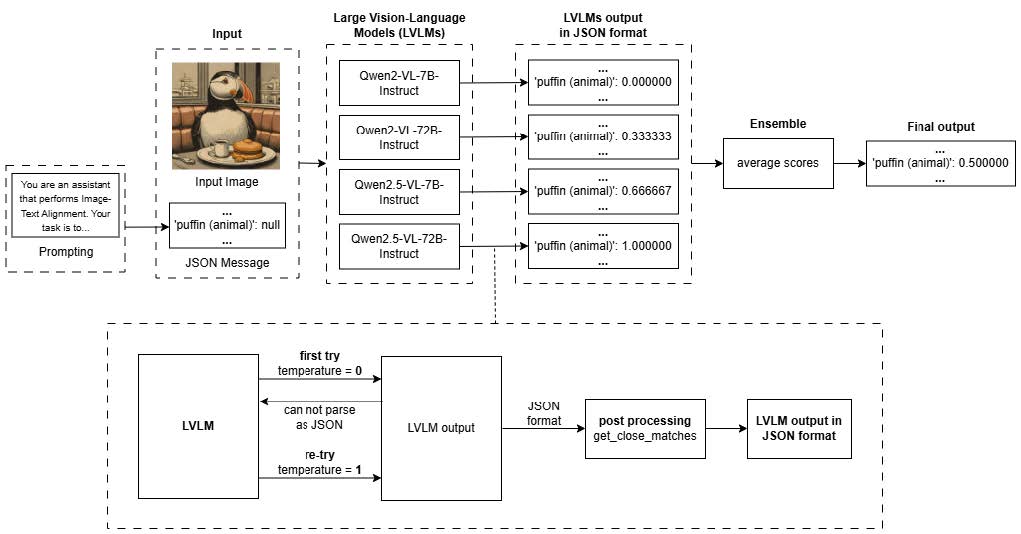}
    \caption{Overview of team HCMUS proposed method.}
    \label{fig:hcmus}
\end{figure*}
\subsubsection{Evalthon}
\label{evalthon}

Team Evalthon~\cite{zhang2025tokenfocusvqae} win second place in the image-text alignment track.
They train multiple Large Vision Language Models(LVLMs) and integrate them using an XGBoost model.
The baselines are Qwen2.5-VL-7B,InternVL2.5-26B and InternVL2.5-38B\cite{internvl2.5}. 

As depicted in \Cref{fig:evalthon}, the training procedure consists of three stages.
Initially, they partition the dataset into five folds,allocating 80\% for training and 20\% for validation within each fold.
The guiding principle is to eliminate duplicate prompts, while allowing image generation models to overlap across folds, aligning with the testing data distribution.
Subsequently, they utilize a combination of image and text, integrating certain statistical features directly as textual inputs.
During the training phase, a low rank\cite{hu2022lora} training technique is employed.
In the training step, they directly extract the logits corresponding to token positions from the hidden state, proceeding to apply a weighted summation leveraging the Mean Squared Error(MSE) loss to fit the model against the label. 

When testing, they utilize various models to predict the test set by deploying checkpoints derived from training across different folds.
Finally, they employ XGBoost\cite{chen2016xgboost} to integrate the predicted scores with selected statistical features, consolidating them into the final score.

\subsubsection{HCMUS}
\label{hcmus}
\begin{figure*}[!t]
    \centering
    \includegraphics[width=\linewidth]{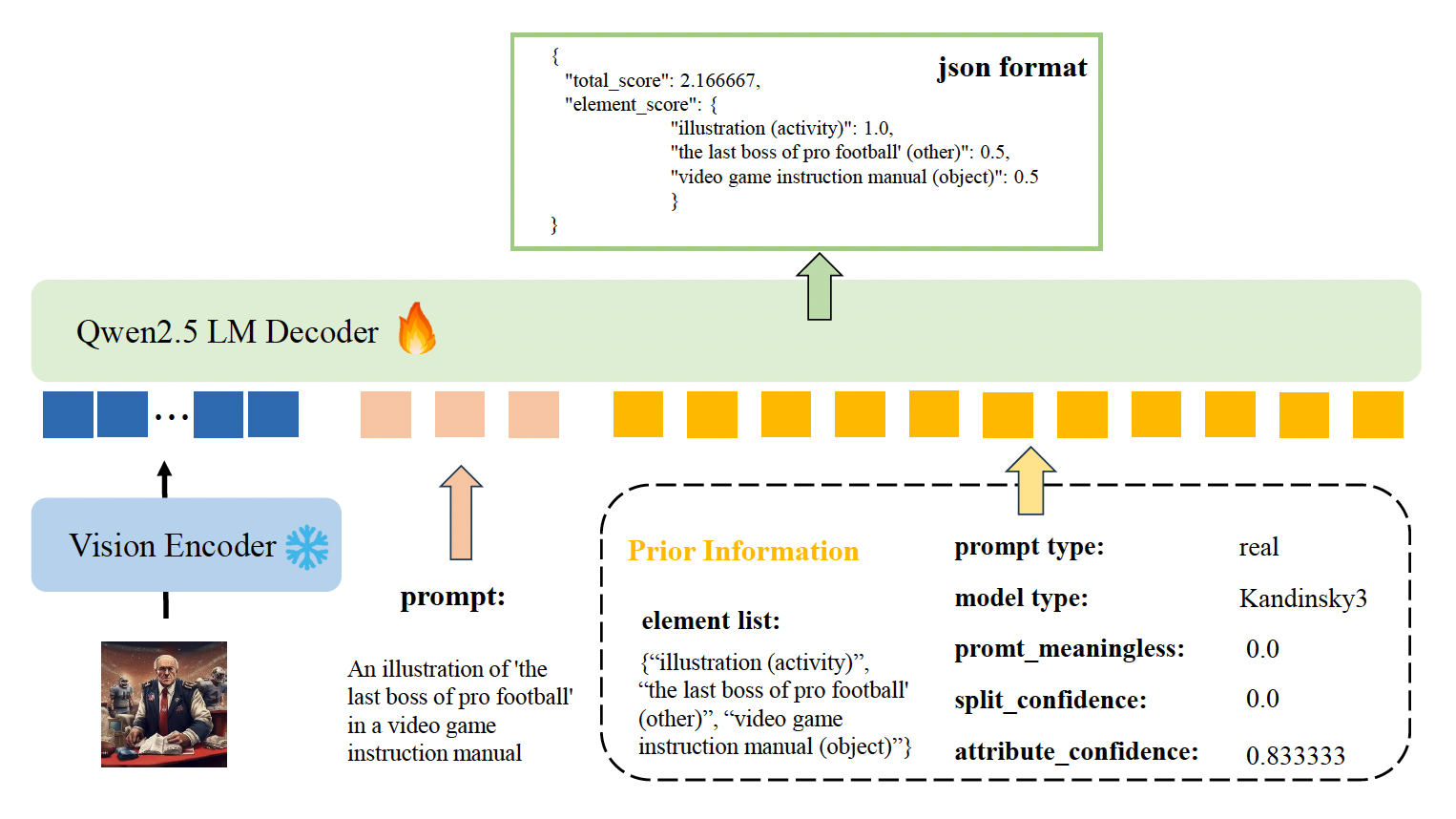}
    \caption{Overview of team MICV proposed method.}
    \label{fig:micv}
\end{figure*}
Team HCMUS wins third place in the image-text alignment track. 
They propose an approach focusing on fine-tuning the Qwen2-VL and Qwen2.5-VL models using Low-Rank Adaptation (LoRA) techniques. 
They also apply ensemble approaches, post-processing, and utilize external datasets to further generalize and achieve better performance, aiming to offer a more refined, human-aligned evaluation framework that can assess image-text alignment with greater precision and sensitivity to fine-grained details.

\Cref{fig:hcmus} illustrates the comprehensive pipeline of their approach. 
Their baseline models are Qwen2-VL-7B-Instruct, Qwen2.5-VL-7B-Instruct, Qwen2-VL-72B-Instruct and Qwen2.5-VL-72B-Instruct. 
For training datasets, they average the element-wise scores in original dataset to produce a single score per field. The processed dataset is formatted into instruction-based input-output pairs suitable for VLM training. 
They also incorporate external datasets, TIFA\cite{hu2023tifa} and GenAIBench\cite{li2024genai}, to enhance model’s performance in text-to-image generation quality assessment. 

They first train Qwen2.5-VL-7B-Instruct with formatted original dataset to directly output the results in a JSON structure.
For TIFA and GenAIBench, they filter out incomplete or invalid samples from these two extra datasets to ensure high-quality data. 
Items with non-JSON-parsable or null-valued outputs are discarded. 
Then they use this trained model to generate pseudo labels and merge the pseudo-labeled datasets with original EvalMuse dataset for training Qwen2.5-VL-72B-Instruct. 
They denote this dataset as External, while they also construct External Private dataset by removing the type attribute. 
Specifically, they train 6 models and simply average their results to obtain final scores. 
The trained models are as following: Qwen2-VL-7B-Instruct(Original), Qwen2.5-VL-7B-Instruct(Original), Qwen2-VL-72B-Instruct(Original), Qwen2.5-VL-72B-Instruct(Original), Qwen2.5-VL-72B-Instruct(External), Qwen2.5-VL-72B-Instruct(External Private).

\subsubsection{MICV}
\label{micv}

Team MICV propose a prior information-guided multimodal approach.
The main focus is to learn the generation characteristics of different generative models from the training data, the alignment level of different elements in various prompts, and to consider the multi-annotator process in the quality assessment task. 
They utilize Qwen2.5-VL to model and predict the global alignment score and local element alignment score.

By leveraging the structure of the original data, they construct global and local element question-answer templates. 
Based on the generative model information from the training data and the relevant information from human annotations, their model predicts the data distribution. 
\Cref{fig:micv} shows the overview of their methods.

\subsubsection{SJTU-MMLab}
\label{sjtummlab}
\label{sjtumm}
\begin{figure*}[!t]
    \centering
    \includegraphics[width=\linewidth]{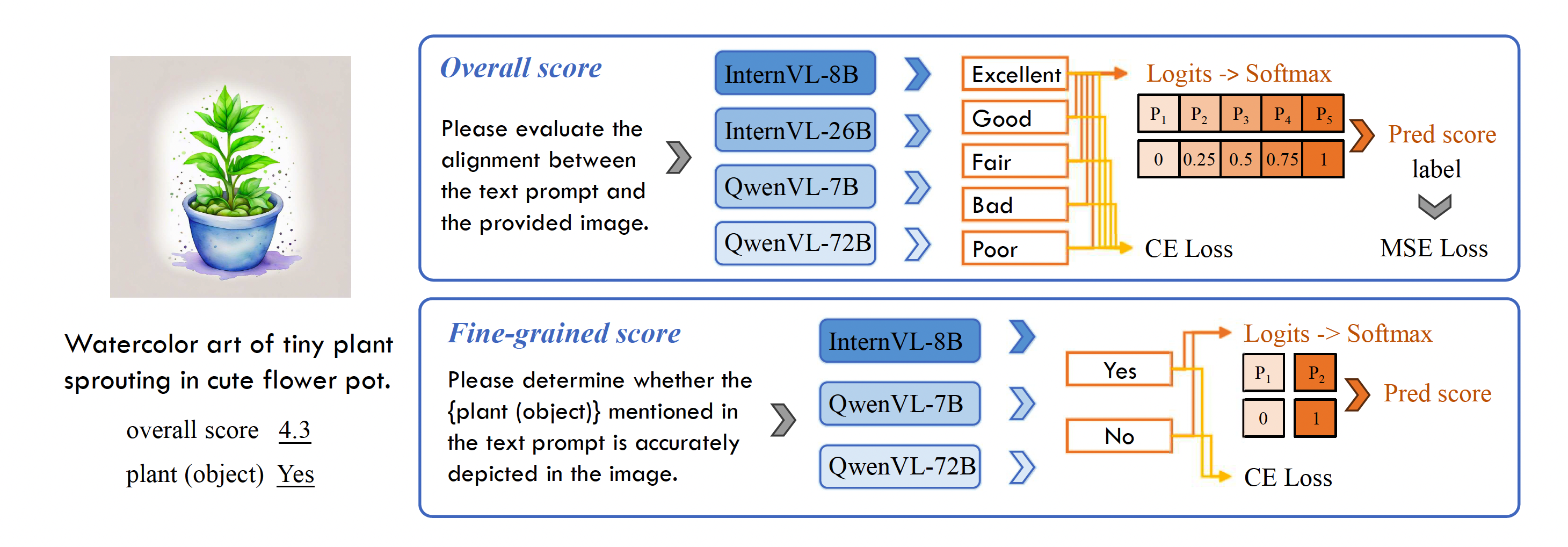}
    \caption{Overview of team SJTU-MMLab proposed method.}
    \label{fig:sjtummlab}
\end{figure*}

Team SJTU-MMLab propose a method similar to Q-Align\cite{wu2023qalign}.
For the overall alignment score, they fine-tune four powerful Multimodal Large Language Models (MLLMs)—InternVL-8B, InternVL-26B, QwenVL-7B, and QwenVL-72B leveraging supervised fine-tuning (SFT) and Low-Rank Adaptation (LoRA) to optimize the vision-language components of the models. \Cref{fig:sjtummlab} shows the overview of their methods. 
Each model outputs a quality score in five categories: Excellent, Good, Fair, Poor, and Bad. 
The corresponding probabilities for each category are weighted according to predefined criteria, and the final overall alignment score is calculated by averaging the weighted probabilities across all models. 
To train the models, they use a combination of Cross-Entropy Loss (CE loss) and Mean Squared Error (MSE loss), ensuring that the models classify the alignment correctly and predict the quality scores with high accuracy. 
For the fine-grained alignment score, they fine-tune three MLLMs—InternVL-8B, QwenVL-7B, and QwenVL-72B—using the same SFT and LoRA approach.
The models perform binary classification, outputting either ”Yes” or ”No” to indicate the presence of the element in the image. 
The final fine-grained alignment score is derived by averaging the probabilities of ”Yes/No” decisions from all models. 
The novelty lies in the combination of MSE loss and CE loss, which allows for both precise classification of alignment quality and fine-tuning of alignment score values, leading to more accurate and reliable measurements. 
Additionally, designing custom prompts for each element category enables the model to focus on the unique characteristics of each element, enhancing its ability to reflect specific features in the generated image.

\subsubsection{SJTUMM}
\begin{figure*}[!t]
    \centering
    \includegraphics[width=\linewidth]{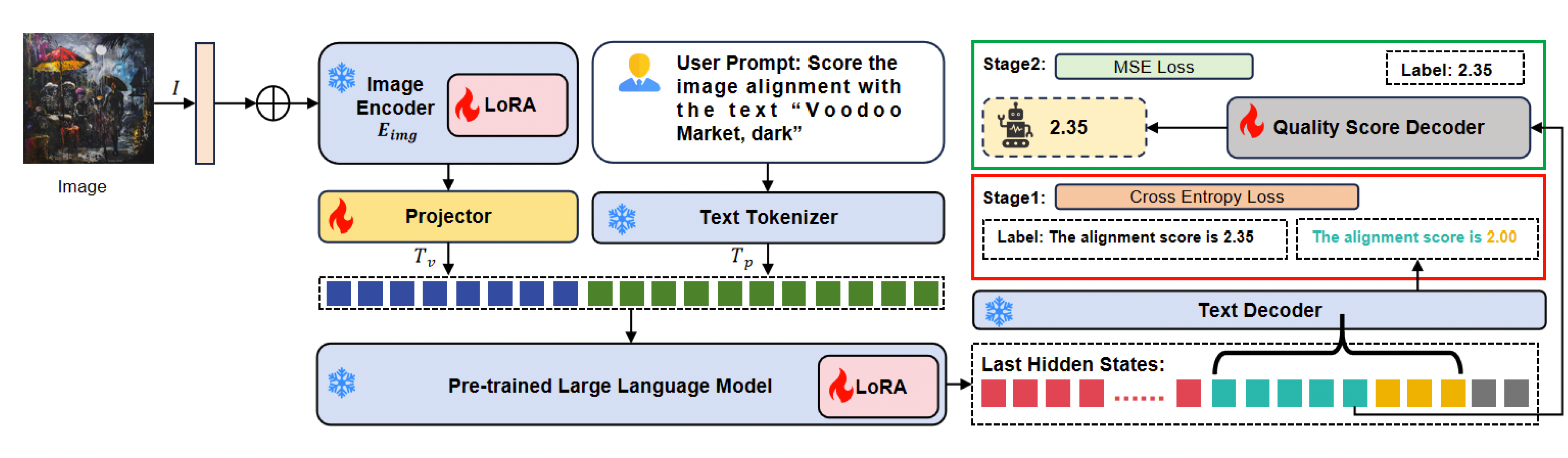}
    \caption{Overview of team SJTUMM proposed method.}
    \label{fig:sjtumm}
\end{figure*}
Team SJTUMM propose a two-stage training stage for Multimodal Large Language Models.
\Cref{fig:sjtumm} is the overview of training method. 
In the first training stage, they use cross-entropy loss to adapt MLLMs to a fixed output format like 'The alignment score is {a floating number}.' 
In the second training stage, the last hidden state representing the token just before the score is decoded through a quality score decoder(not the lm head), with the score number and mean squared error loss used for training. 
They change the LLM backbone as InternVL2.5, Qwen2.5VL and DeepSeekVL2\cite{deepseekvl2} and average the total scores. 
The element scores are obtained by that if one of the three models determines that there is the element, then this element score is one.

\subsubsection{WT}
\label{wt}
Team WT adopt the methods proposed in Q-Align and DeQA-Score\cite{deqascore}. They choose InternVL2.5-8B as baseline model. 
They design tailored instructions, with responses for image-text alignment categorized into bad, poor, fair, good, and excellent, while element presence is evaluated with a yes/no response. 
To ensure robust and accurate scoring, they employ a dual-method strategy:
\begin{itemize}
    \item \textbf{Weighted Probability-Based Scoring\cite{wu2023qalign}}. 
    Image-text alignment scores are computed by weighting the probabilities of the five response categories. Element presence scqres are derived from the probability of a "yes" response.
    \item \textbf{Score Distribution Alignment via KL Divergence\cite{deqascore}}. 
    Recognizing that discrete textual responses do not naturally translate into continuous scores, they introduce a score distribution modeling technique. 
    They treat the mean opinion score (MOS) as a Gaussian distribution, computing probabilities across the five response levels. 
    By applying KL divergence, they align the predicted probability distribution with the true distribution, ensuring more precise and reliable scoring. 
    KL divergence is also applied to element presence evaluation, aligning the probabilities of "yes" and "no" with true probabilities.
\end{itemize} 
They model the MOS as a Gaussian distribution and use KL divergence to align the predicted distribution with the actual one, improving the accuracy of score prediction:
\begin{itemize}
\item \textbf{Interval probability:} They divided the MOS into 5 intervals and used the mean and variance of each MOS label to estimate the probability of it belonging to each interval. 
The probability calculation formula is as follows:
\begin{equation}
p_i = \int_{i-0.5}^{i + 0.5} f(x) \, dx, \quad i \in \{1, 2, 3, 4, 5\}
\end{equation}
where \( f(x) \) is the Gaussian probability density function for each MOS.
\item \textbf{Probability adjustment:} The sum of the five interval probabilities modeled in this way does not equal to 1, leading to MOS shift. 
Therefore, they apply a linear transformation to ensure the accuracy of MOS. The formula is as follows:
\begin{equation}
p_i^{\text{new}} = \alpha p_i + \beta 
\end{equation}
By solving the following equations, they obtain the values of \( \alpha \) and \( \beta \):
\begin{equation}
\begin{cases}
\sum_{i=1}^{5} p_i^{\text{new}} = 1, \\
\sum_{i=1}^{5} p_i^{\text{new}} i = MOS.
\end{cases} 
\end{equation}
\item \textbf{Loss function:} They use KL divergence to align the predicted probabilities of the five tokens from the llm with the interval probabilities above:
\begin{equation}
L_{kl} = \sum_{i=1}^{5} p_i^{\text{new}} \log\left(\frac{p_i^{\text{pred}}}{p_i^{\text{new}}}\right) 
\end{equation}
\end{itemize}

During the testing phase, they fused the results from three checkpoints.
Specifically, during training, they saved a checkpoint every 800 steps and selected the three best-performing checkpoints on the validation set (3200 steps, 4000 steps, and 4800 steps) for testing on the test set. 
The final results were obtained by averaging their outputs.

\subsubsection{YAG}
\label{yag}
\begin{figure*}
    \centering
    \includegraphics[width=\linewidth]{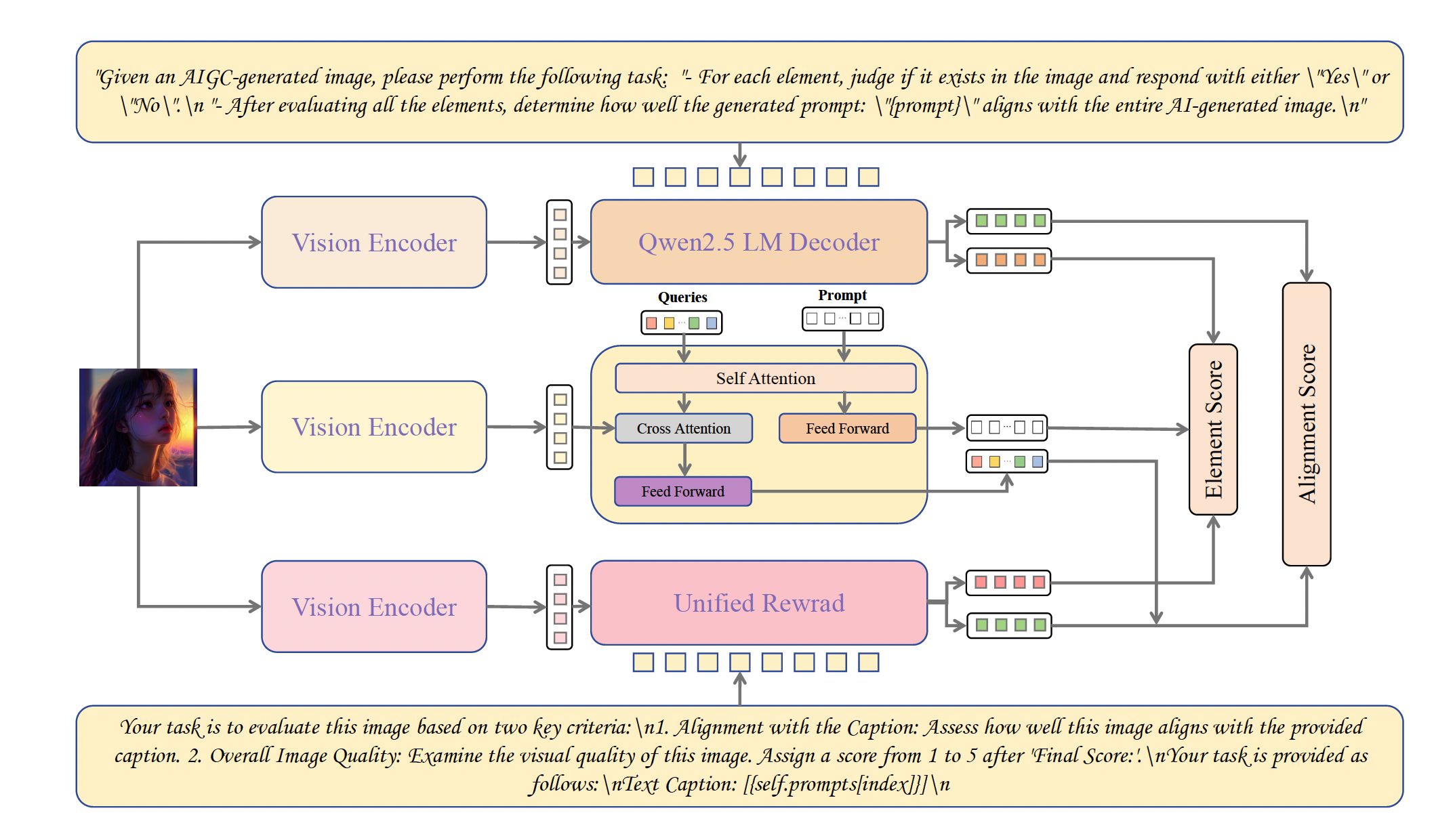}
    \caption{Overview of team YAG proposed method.}
    \label{fig:yag}
\end{figure*}
They propose a multi-granularity quality assessment framework for AI-generated content (AIGC) by leveraging the robust multimodal capabilities of foundation models such as Qwen2.5-VL and LLAVA-onevision.\cite{li2024llava} \Cref{fig:yag} shows the overview of their methods.
To evaluate text-image alignment and quality, they design model-specific question answering templates that guide the models to assess both element-level details and overall coherence.
For Qwen, the prompt instructs the model to evaluate the presence of predefined elements in the image with binary
responses and judge the alignment between the generation prompt and the image on a 1.0–5.0 scale.
For LLAVA, following a Unified Reward\cite{wang2025unified} inspired structure, the model evaluates alignment with the caption, visual quality, and extracts element presence scores (0/1) using structured output formats.
To enhance cross-modal interactions, they integrate the Qformer architecture from the FGA BLIP2\cite{han2024evalmuse} framework, enabling joint analysis of element-level and holistic scores. 
Critically, to address potential omissions in element detection, they adopt an ensemble strategy where an element is marked as ”present” if identified by any model, ensuring comprehensive coverage while mitigating individual model biases. 
This approach combines structured scoring with robust element detection to achieve reliable AIGC quality assessment.

Additionally, they employ an ensemble of three distinct multi-modal models: FGA-BLIP2, Qwen2.5-VL (3B and 7B variants), and Unified Reward (7B version). 
Specifically, Qwen Variants are the untrained base Qwen (serving as a control reference),
a LoRA-adapted Qwen (for parameter-efficient fine-tuning) and a fully
fine-tuned Qwen (with full parameter optimization).
The outputs from these Qwen variants are aggregated with the Inference results from Unified Reward (trained on the EvalMuse-40k dataset), and trained FGA-BLIP2 outputs, to form the final detection ensemble.

\subsubsection{SPRank}
\label{sprank}
Team SPRank propose Qwen-Assisted Image-Text Alignment Scoring. 
They leverages a finetuned Qwen2.5-VL-7B-Instruct model to enhance both alignment and element scoring for image-text assessment. 
First, they finetune on the given data question-answer pairs and then generate custom element questions for further specialized fine-tuning. 
For alignment scoring, they use specifically designed prompts to guide the model’s evaluation. 
Finally, they combine the fine-tuned model scores with baseline scores through weighted summation, optimizing both element existence judgment and image-text alignment assessment.

During testing, they ensemble the results from fine-tuned model and the baseline method using weighted summation to produce the final results:
\begin{equation}
    \begin{aligned}
        S_a = 0.5 \times S_{a_{\text{baseline}}} + 0.5 \times S_{a_{\text{qwen}}} \\
        S_e = 0.7 \times S_{e_{\text{baseline}}} + 0.3 \times S_{e_{\text{qwen}}}
    \end{aligned}
\end{equation}
\( S_a \) denotes the alignment score, while \( S_e \) represents the score generated by the fine-tuned Qwen2.5 model.

\subsubsection{AIIG}
\label{aiig}
\begin{figure}
    \centering
    \includegraphics[width=\linewidth]{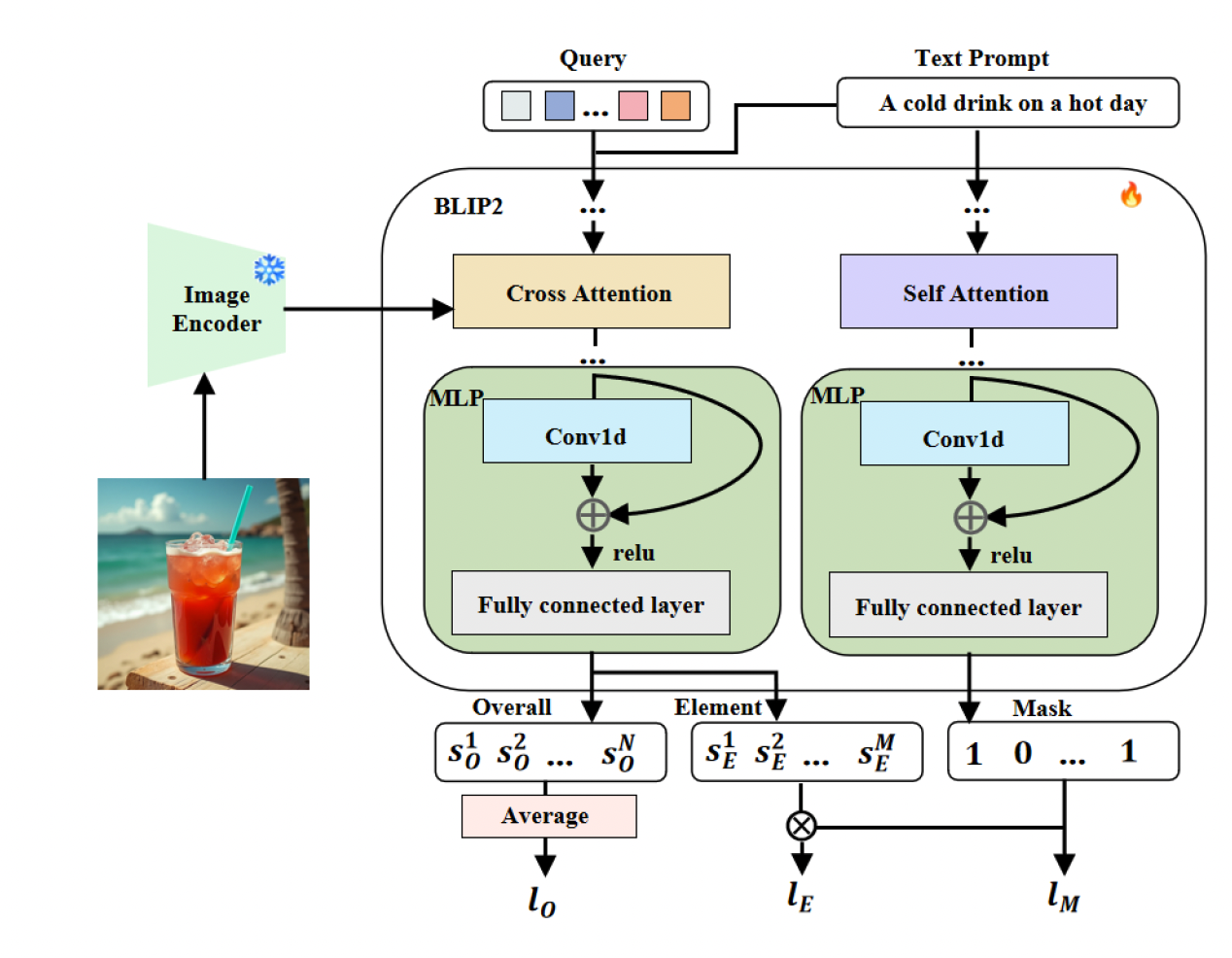}
    \caption{Overview of team AIIG proposed method.}
    \label{fig:aiig}
\end{figure}
Team AIIG selected the model FGA-BLIP2 as the baseline. 
FGA-BLIP2 enables fine-grained alignment evaluation by combining training of overall and element alignment scores. 
They use the ITM setup of BLIP2\cite{li2023blip2} to concatenate the query and embedded text, and
then cross-focus with the image. 
The final alignment score is obtained by a two-class linear classifier, where the query sections are averaged to produce an overall alignment score, while the text sections in each corresponding position provide an alignment score for a specific element. 
Text prompts are first fed to the self-attention layer to extract text features, which are then passed to the MLP. 
The MLP module captures the local context pattern in the sequence through the Conv1d layer, and maintains stable gradient flow while deepening the network depth with residual connections. 
The multi-scale features of its output are then fused with the fully connected layer. 
The final MLP predicts the
mask that represents the validity of each text tag.

As show in \Cref{fig:aiig}, \( N \) and \( M \) are the length of the query and text prompt, respectively.
\( S_O \) and \( S_E \) respectively represent the overall and split element scores for image-text alignment.
\( l_O \) represents the loss between the predicted overall alignment score and the manual annotation, \( l_E \) represents the loss between the predicted element alignment score and the manual fine-grained annotation, and \( l_M \) represents the loss between the predicted valid text annotation and the real element.
In addition, they use variance to weight the loss function.
The greater the variance of the image-text pair, the higher the loss. 
The final loss objective function is shown in formula (\ref{eq:1}). 
Where \( \sigma \) is the variance of the overall alignment score of different images generated under the same text prompt. 
The weight parameters \( \alpha \) and \( \beta \) are set to \( 0.1 \). 
This method makes the model pay more attention to the samples with large alignment score difference in the training process, thus improving the accuracy and robustness of the evaluation.

\begin{equation}
\label{eq:1}
L = e^{\sigma} \times (l_O + \alpha l_E + \beta l_M)
\end{equation}

\subsubsection{Joe1007}
\label{joe1007}
Team Joe1007 enhance FGA-BLIP2 by using a multi-component loss function that combines the original ITM (Image-Text Matching) loss with three specialized components. 
During training, the model processes image-text pairs through the visual encoder and Qformer\cite{li2022blip} architecture, generating embeddings and scores for both the overall match and individual elements within the prompt. The training process utilizes:
\begin{itemize}
    \item A base loss derived from the original FGA-BLIP2 model (score difference,
    token score, and mask prediction)
    \item Distribution matching loss that guides the model toward generating
    predictions with similar distribution characteristics to the validation set
    \item Element type adaptive weighting based on statistical properties of different
    element types
    \item Contrastive learning to ensure elements of the same type have similar
    feature representations
\end{itemize}
Training uses a linear warmup followed by cosine learning rate scheduling,
with image augmentation techniques like random resizing and cropping.
The model leverages pre-trained weights from BLIP2 and fine-tunes on the
fine-grained evaluation dataset while keeping the vision encoder frozen to
prevent overfitting.

\subsubsection{iCOST}
\label{icost}
Team iCOST utilizes the training policy used in DeepSeek-R1\cite{guo2025deepseek} in this challenge scenario. 
They choose two baseline models: Qwen2.5-VL-72B and Qwen2.5-VL-7B. First, they transform the given dataset into image-text conversation.
Second, they pick some of them according to the attribute confidence, which higher confidence may reflect higher quality. 
Then they let Qwen2.5-VL-7B output the evaluation scores and fine-tune it to make the output consistent with the ground truth by SFT. 
Then they pick other high-quality image-text to continue to fine-tune the model by GRPO.

\subsection{Structure Distortion Detection}
\begin{figure*}[!t]
    \centering
    \includegraphics[width=\linewidth]{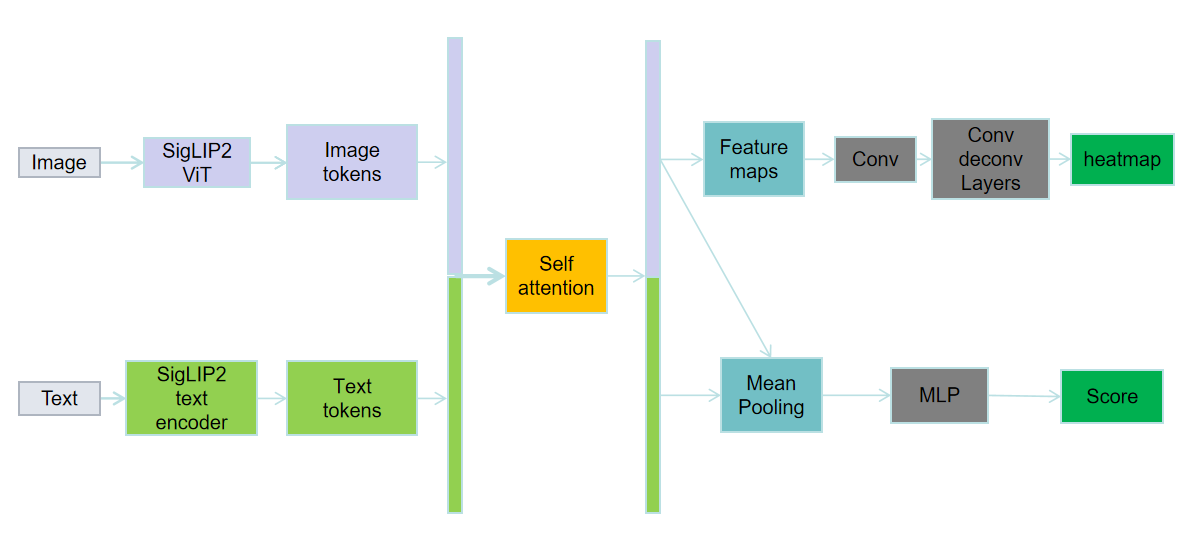}
    \caption{Overview of team out of memory proposed method.}
    \label{fig:outofmemory}
\end{figure*}
\subsubsection{HNU-VPAI}
\label{hunvpai}
Team HNU-VPAI wins the championship in the structure distortion detection track.This team used separate models for two subtasks. 
They use an Image Quality Assessment (IQA) model to predict quality scores and an instance segmentation model to identify structural issue regions in the image.

For the image quality assessment sub-task, this team adopt a CNN-Transformer hybrid architecture to leverage the strengths of both models in detecting structural distortions.
Convolutional Neural Networks (CNNs) excel at capturing fine-grained, local features, making them particularly effective at identifying localized distortions in the image, such as misalignments or local texture anomalies.
On the other hand, Transformers are highly capable of capturing long-range dependencies and global context, which allows them to understand the overall structure of the generated image and detect distortions that might not be immediately apparent in local regions but affect the image's global coherence. 
By combining CNNs and Transformers, this approach benefits from the detailed local feature extraction of CNNs and the global context modeling of Transformers. 
This hybrid architecture is particularly effective for structure distortion detection, as it enables precise identification of both local and global distortions, leading to more accurate and robust quality assessments of generated images. 
Inspired by LoDa~\cite{xu2024boosting}, this team first employ a CNN to extract local distortion features from input images, and then inject the CNN-extracted multi-scale local distortion features into the ViT using a cross-attention mechanism. 
This allows the ViT to focus on the distortion-related features while maintaining its strength in capturing global context. 
By querying relevant multi-scale features from the CNN and fusing them with the ViT’s image tokens, they ensure that both local and global distortions are captured simultaneously. 
To further optimize efficiency and reduce computational overhead, they follow LoDa to down-project the high-dimensional ViT tokens and multi-scale features into a smaller dimension before performing cross-attention. 
This ensures that the model remains computationally efficient while leveraging the rich information from both the CNN and ViT. 
It is important to note that both the CNN and ViT are pretrained models, and their parameters remain frozen during training. 
They only update the parameters of the lightweight cross-attention module. 
This approach retains the rich knowledge embedded in the pretrained models while focusing the training on the task-specific components, ensuring efficient adaptation with minimal computational overhead.
They conducted extensive experiments with various variants of pretrained CNN and ViT models and found that a simple ResNet architecture performed well for the CNN, while the CLIP and DINOv2 pretrained models yielded the best results for the ViT.

For the structural distortion detection sub-task, this team utilize the state-of-the-art instance segmentation model (e.g., Co-DETR~\cite{zong2023detrs}) to identify regions with structural distortions. 
Even though the baseline method frames this task as a semantic segmentation problem, they found that treating it as an instance segmentation task leads to better results. 
Co-DETR is an advanced instance segmentation model that enhances the performance of DETR-based detectors by incorporating a collaborative hybrid assignments training scheme. 
This approach utilizes versatile one-to-many label assignments, such as ATSS and Faster RCNN, to enrich the supervision provided to the encoder's output, thereby improving its discriminative capabilities.
Additionally, Co-DETR introduces customized positive queries by extracting positive coordinates from auxiliary heads, which enhances the training efficiency of the decoder. 
During inference, these auxiliary heads are discarded, ensuring that the method does not introduce extra parameters or computational costs to the original detector. 
In the context of structural distortion detection, Co-DETR's architecture is particularly effective. Its encoder-decoder structure, combined with the collaborative hybrid assignments training scheme, allows for the precise identification of regions with structural distortions. 
By treating this task as an instance segmentation problem, Co-DETR can distinguish individual distorted regions, leading to more accurate detection compared to semantic segmentation approaches.

\subsubsection{OPDAI}
\label{opdai}
\begin{figure*}[!t]
    \centering
    \includegraphics[width=\linewidth]{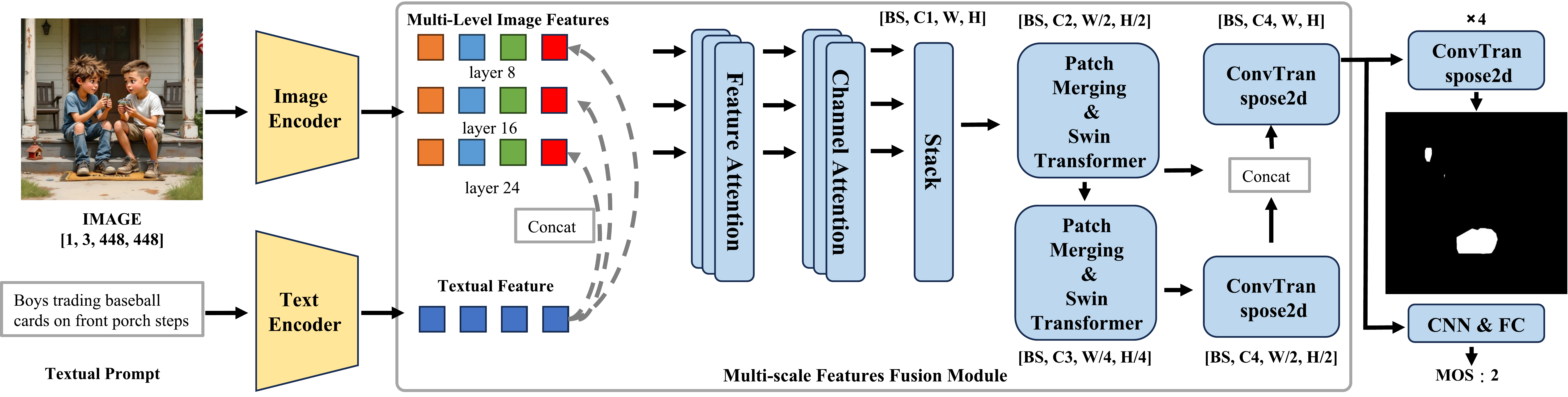}
    \caption{Overview of team I$^2$ Group proposed method.}
    \label{fig:i2group}
\end{figure*}
Team OPDAI propose a multi-task architecture leveraging Florence2's foundation model~\cite{xiao2024florence} capabilities for simultaneous structural distortion localization (heatmap prediction) and severity estimation (implausibility scoring).  The network employs: 
\begin{itemize}
\item Florence2-large encoder (frozen weights initialized from FLD-5B pretraining)
\item U-Net style decoder~\cite{ronneberger2015u}  for heatmap prediction, optimized using a combination of Mean Squared Error (MSE) and Dice-IoU losses.
\item Multi-layer perceptron head for regression scoring, trained with a composite loss comprising MSE,PLCC loss and SRCC loss.
\end{itemize}
They use a two-stage training strategy, heads-only training for 20 epochs with and lr=1e-4, then full fine-tuning training for another 50 epochs with lr=1e-5 for backbone and 1e-4 for heads. There are different losses for the two tasks:
\begin{itemize}
  \item Heatmap: 
  \begin{equation}
  \mathcal{L}_{hm} = \text{MSE} + (1 - \text{Dice-IOU})
  \end{equation}
  \item Score: 
  \begin{equation}
      \mathcal{L}_{score} = \text{MSE} + 0.3(1-\text{PLCC}) + (1-\text{SRCC})
  \end{equation}
\end{itemize}

\subsubsection{MICV}
\label{micv_struct}

Team MICV propose SGL-SDD, a CLIP-based model tailored for detecting the structure distortion in the images generated by T2I models. 
To reduce the label noise introduced by annotators, they adopt the study group learning strategy for the detection learning. 

This Team observe that it would be better to divide structure distortion detection task into two models for learning, and design two models, one model for the score learning and one model for the heatmap learning. For each learning, they adopt the strategy of Study Learning Group(SLG~\cite{zhou2021study}), to reduce the label noise in the training set.
Specifically, for each learning, they randomly and averagely split the whole training set $(X, Y)$ into $K$ subsets $\{(X_k, Y_{k})\}$, where $k \in [1, K]$. 
They train totally $K$ models $\{M_k\}$. For each model $M_k$, they train it on $(X \setminus X_k, Y \setminus Y_{k})$. 
After training the model set, they infer the estimated score or heatmap label $\tilde{Y}_{k}$ of $X_k$ as the pseudo label based on the model $M_k$. 
Finally, they take the average of the true label and the pseudo label as the label for model training. 
Besides, they also adopt the strategy of prompt rewriting to make text data augmentation based on the GPT-4o.
They adopt the pre-trained model SigLip2~\cite{tschannen2025siglip} in the model. 
The training time for each model on 8 A100 GPUs is 17 hours. 
The parameters on the models are 300+ millon.

They adopt the strategy of test-time augmentation in the testing. 
For each test sample, they conduct the model prediction six times for different prompts generated by prompt rewriting, and take the average of these results as the final prediction result.

\subsubsection{out of memory}
\label{outofmemory}

Team out of memory optimize baseline~\cite{han2024evalmuse, liang2024rich} architecture of structure distortion detection track with several key modifications and optimizations to improve performance. 
The model is designed to jointly predict heatmaps and scores from images and text.

They employ siglip2-so400m-patch16-512~\cite{tschannen2025siglip}  as the backbone of model architecture. 
The last hidden states from both the visual and textual encoders are concatenated along the token dimension, yielding a unified multimodal representation that seamlessly integrates information from both modalities. 
They perform two key optimizations on self-attention module, one is using the xformers library for memory-efficient attention computation, another is replacing the traditional LayerNorm with RMSNorm: A normalization layer that stabilizes training by normalizing activations. 
The heatmap predictor also replaces LayerNorm with RMSNorm, along with adjustments to the normalization dimension for improved efficiency. 
They simplify the score prediction branch by replacing the baseline's convolutional and MLP-based approach with a more straightforward design. 
The new architecture computes a global score from the refined multimodal embeddings by Mean Pooling and Fully Connected Layers. 
This streamlined approach effectively summarizes multimodal features into a single confidence score while reducing complexity.

They utilize the AdamW Legacy optimizer with an initial learning rate of $2.1 \times 10^{-5}$, combined with the Cosine Annealing LR learning rate scheduler to ensure stable convergence.
To accelerate training and optimize memory usage, they employ automatic mixed precision (AMP).
Additionally, the optimizer is configured with a weight decay of 0.01 and the `caution` flag set to `True`. 
The `caution` mechanism skips updates for certain parameters when the update direction conflicts with the current gradient direction, mitigating potential instability caused by noisy gradients and ensuring more robust training under noisy conditions. 
The loss function is based on Mean Squared Error (MSE) for both components of objective. Specifically, the total loss is computed as:
\begin{equation}
    \text{total\_loss} = 16 \cdot \text{loss\_heatmap} + 4 \cdot \text{loss\_score}
\end{equation}
where $\text{loss\_heatmap}$ corresponds to the MSE loss for the heatmap prediction task, and $\text{loss\_score}$ corresponds to the MSE loss for the score prediction task.

The training process involves 3,000 iterations, with a weight decay of 0.01, a batch size of 1, and gradient accumulation over 32 steps. 
All experiments are performed on a single NVIDIA H100 GPU equipped with 80GB of memory. 
The training dataset comprises all data provided by the competition organizers, and the overall training process takes approximately 5 hours to complete.

\subsubsection{I$^2$ Group}
\label{i2group}
\begin{figure*}
	\centering
	\begin{subfigure}{0.49\linewidth}
		\centering
		\includegraphics[width=\linewidth]{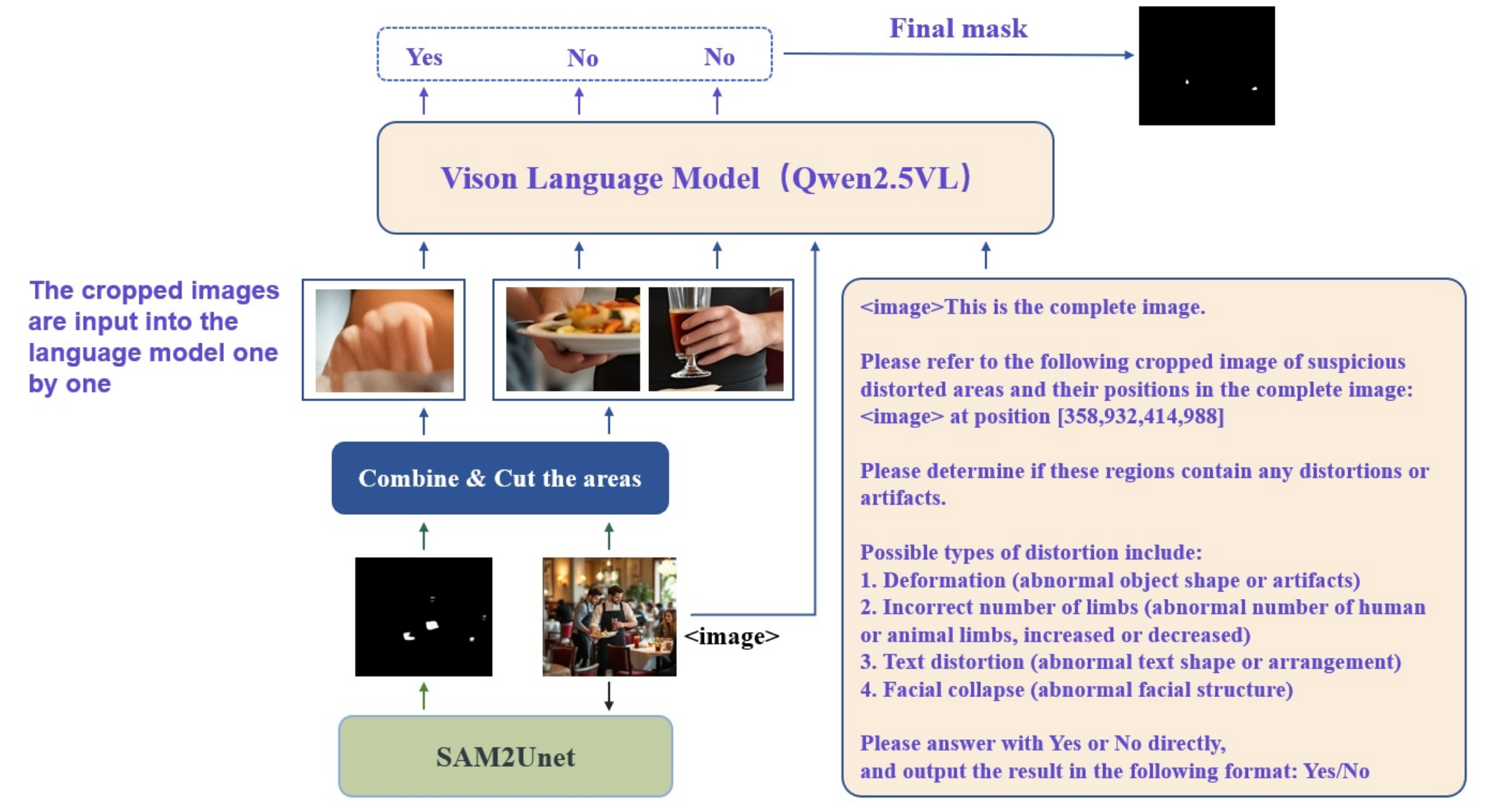}
		\caption{Score Prediction Pipeline}
		\label{bruteforcewins01}
	\end{subfigure}
	\centering
	\begin{subfigure}{0.49\linewidth}
		\centering
		\includegraphics[width=\linewidth]{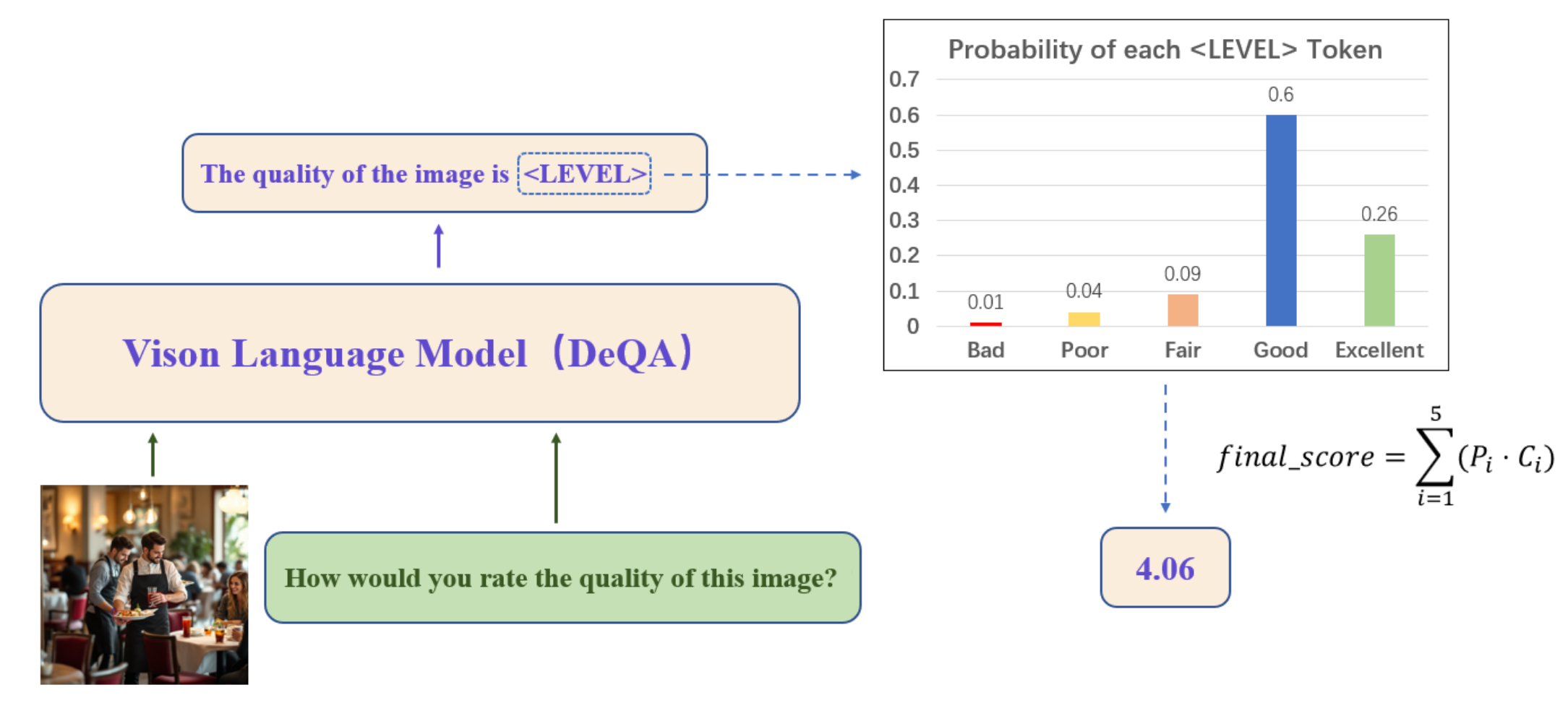}
		\caption{Heatmap Prediction Pipeline}
		\label{bruteforcewins02}
	\end{subfigure}
	\centering
	\caption{Overview of team Brute Force Wins proposed method.}
	\label{bruteforcewins}
\end{figure*}
Team I$^2$ Group propose the MF$^2$M-IQA framework, which leverages multi-level and multi-scale image and prompt features to achieve accurate quality assessment and structure distortion detection for AIGC images.

MF$^2$M-IQA mainly consists of three components: 1) Image and text encoders that extract multi-level visual and textual features; 2) a Multi-scale Features Fusion Module (MF$^2$M) designed to extract multi-scale fusion features by integrating these multi-level visual and textual features; and 3) decoders tailored for downstream tasks—specifically, upsampling (ConvTranspose2D) decoders for predicting distorted region masks, and regression modules composed of CNNs and fully connected layers to estimate corresponding MOS scores. Next, they introduce these three components in details.

Specifically, They adopt AltCLIP \cite{chen2022altclip}, pre-trained on large-scale AIGC images, as image and text encoder. 
Inspired by prior studies \cite{chen2022teacher,chen2024promptiqa,wang2023Hierarchical,yang2022maniqa,chen2024seagull} demonstrating that multi-level image features enhance image assessment performance, and given that ViT extracts detailed information at shallow layers and semantic information at deeper layers, they select features from layers 8, 16, and 24 as our multi-level image representations.

Then, they propose a Multi-scale Features Fusion Module (MF$^2$M) to effectively fuse visual and textual features. 
Specifically, MF$^2$M employs self-attention mechanisms in both spatial and channel dimensions to model the relationships between images and texts. 
The resulting features are then integrated through efficient Swin Transformers and patch merging blocks \cite{liu2021swin} to capture multi-scale semantic, structural, and detailed information. 
These multi-scale fusion features enhance the model's capability for distortion detection and quality assessment.

Lastly, they design task-specific decoders for structure distortion detection and quality assessment. 
For structure distortion detection, four Conv Transposed 2D layers are employed to upsample multi-scale fusion features to the target mask resolution. For quality assessment, inspired by MANIQA \cite{yang2022maniqa}, they adopt a dual-branch approach to predict pixel-wise quality scores and corresponding weights.
The final quality score is computed as the weighted sum of these predicted pixel scores.

The model trained with four RTX 3090 GPUs, keeping both the image and text encoders frozen. 
Due to the GPU devices limitation, they separately train this model on structure distortion detection and quality assessment tasks. 
They repeatably train MF$^2$M-IQA for each task several times, and finally choose four models for masks prediction and three models for score prediction. 
For score prediction, they additionally train one model by replacing the MF$^2$M with the MANIQA decoder part.
The final results are the average of these predictions.

\subsubsection{Brute Force Wins}
\label{bruteforcewins_}
\begin{figure}
    \centering
    \includegraphics[width=\linewidth]{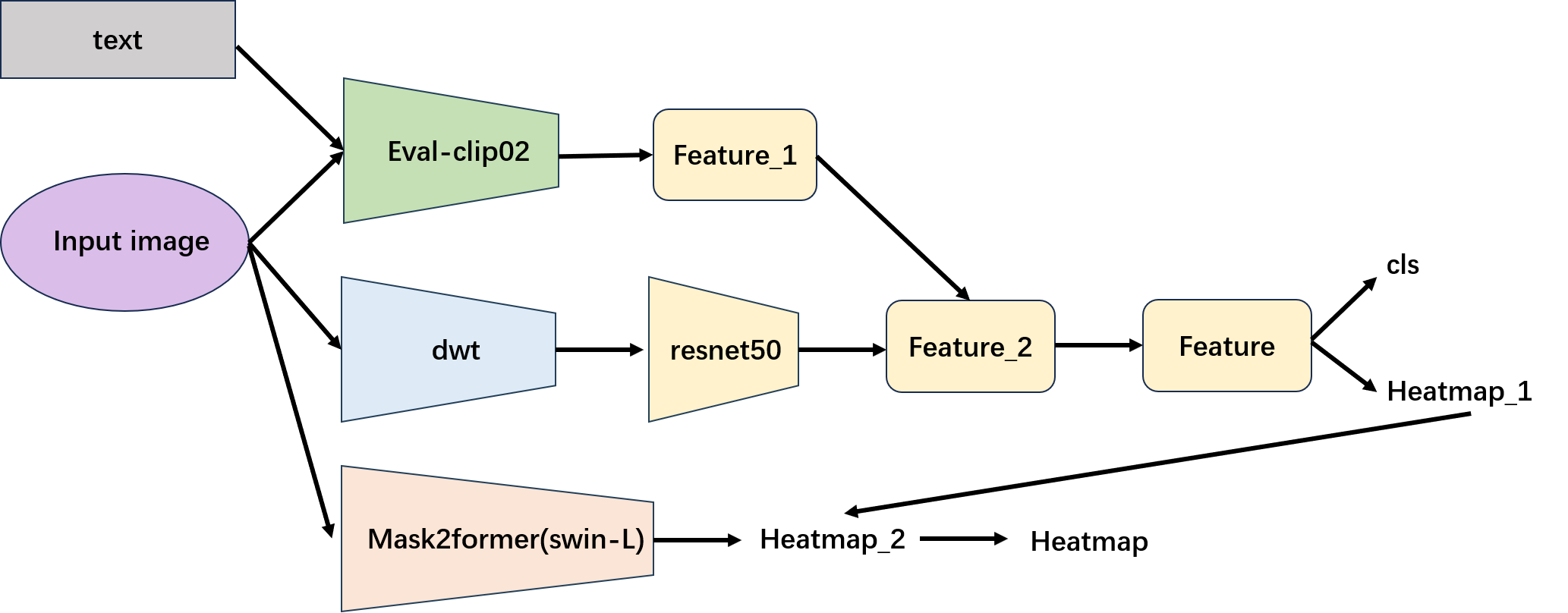}
    \caption{Overview of team Wecan EvalAIG proposed method.}
    \label{fig:wecanevalaig}
\end{figure}
\begin{figure*}
    \centering
    \includegraphics[width=\linewidth]{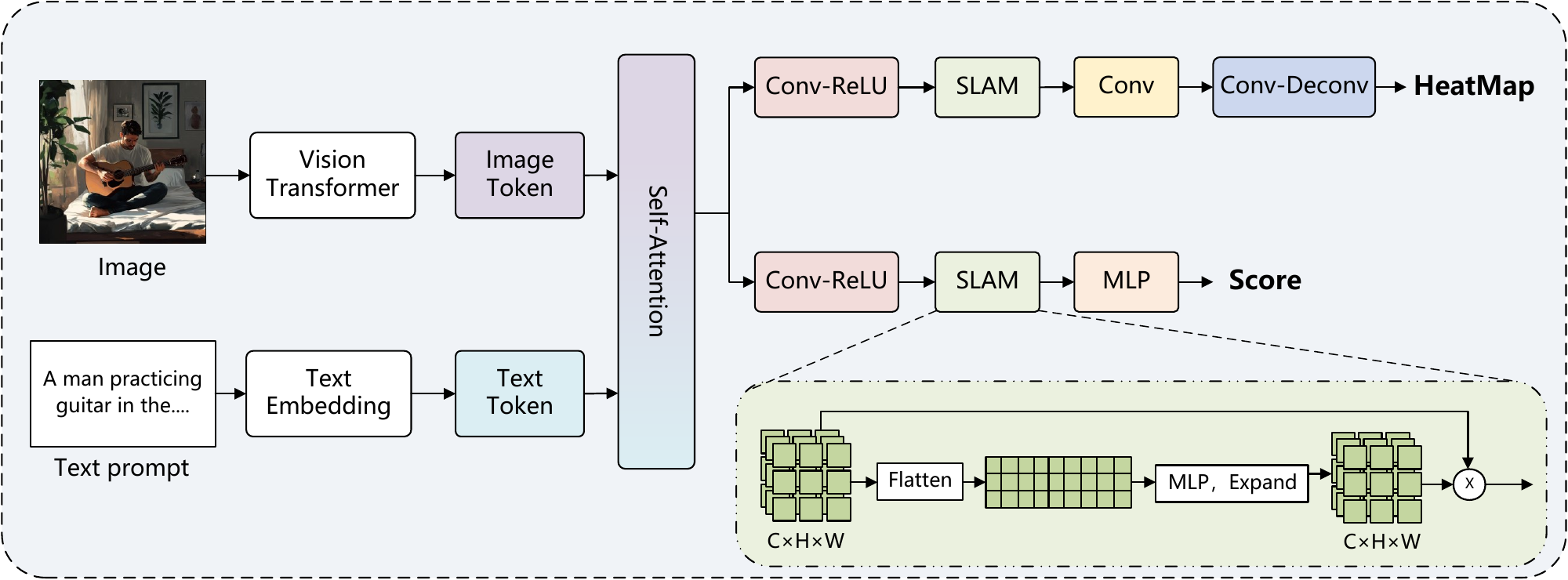}
    \caption{Overview of team Tenryu Badu proposed method.}
    \label{fig:tenryubadu}
\end{figure*}
Brute Force Wins team use separate models for two subtasks.

For the heatmap prediction task, they employ the SAM2Unet\cite{xiong2024sam2} for fine-grained segmentation of distortion regions.
Subsequently, they utilize Qwen2.5VL\cite{bai2025qwen2} to filter and refine the distortion heatmaps predicted by SAM2Unet. 
In detail, they trained a specialized segmentation model to predict and segment distortion regions. 
While the SAM model provides fine-grained region predictions, it lacks advanced semantic understanding, making it unable to determine whether a region is distorted or not. 
For example, the model can detect a human hand in an image but cannot distinguish between a normal hand, a hand with an incorrect number of fingers, or a hand with correctly numbered fingers but misplaced in an incorrect position. 
To address this limitation, semantic understanding capabilities are required. 
To overcome this issue, this team cropped the regions predicted by SAM and sequentially evaluated them using MLLM. 
From a quantitative perspective, both standalone SAM and baseline models suffer from high recall but low precision. 
By introducing MLLM as a discriminator, they achieved a significant improvement in precision while maintaining a relatively high recall, thereby increasing the F1-score.

For the scoring task, Brute Force Wins team adopt a training strategy inspired by DeQA-Score\cite{you2025teaching} and Q-Align\cite{wu2023q} to fine-tune mplug-owl2-llama2\cite{ye2024mplug} for evaluating the structral score of images. 
They prompting the model with "How would you rate the quality of the image?" and allowing it to output a <LEVEL> token. 
The final score was computed as a weighted sum of token probabilities and their corresponding weights. 
With this new method, both training and evaluation times were significantly shorter than those of the baseline, and the final scores marginally outperformed the baseline.

\subsubsection{Wecan EvalAIG}
\label{wecanevalaig}
Team Wecan EvalAIG have designed a two-tier parallel training framework. 
In the first tier, a pre-trained EVA-CLIP model\cite{EVA} is employed for foundational classification and semantic segmentation tasks, while high-frequency features are extracted through a ResNet50 network and Discrete Wavelet Transform (DWT)\cite{yan2024sanity}. 
These features are subsequently fused with those from EVA-CLIP. The second tier implements instance segmentation tasks based on the Mask2Former architecture. 
Ultimately, segmentation results from both tiers are integrated to form comprehensive outcomes.

The model trained on the official training set provided by the competition, without using any additional data. 
The training phase completed 2000 epochs with a batch size of 4*4. Color jitter data augmentation was applied. 
When Testing, they use test time augumentation in mmsegmentation.

\subsubsection{Tenryu Babu}
\label{tenryubadu}

Team Tenryu Badu proposed a unified assessment framework that integrates cross-modal semantic understanding with spatial structural reasoning to enable fine-grained structural quality assessment. 
Specifically, the framework leverages CLIP’s dual encoders to extract visual and textual embeddings from the generated image and its corresponding prompt. 
These embeddings are subsequently fused through a cross-modal self-attention module to identify semantic discrepancies related to structural content. 
The fused representations are then passed into two task-specific branches: 1) a distortion scoring branch, implemented as an MLP-based regression module that predicts scalar scores reflecting structural quality; and 2) a defect localization branch, designed as a U-Net-style decoder with skip connections for pixel-level mask prediction of structural distortions.
A key component of the architecture is the Spatial Layout Attention Module (SLAM), which adaptively reweights channel-wise feature responses based on their spatial activation patterns. 
By flattening feature maps and processing them through a shared MLP, SLAM learns to identify and enhance sensitivity to geometric anomalies like hand distortions or facial collapse. 
Through the joint optimization of structural score regression and defect mask prediction, the proposed framework enables human-aligned evaluation of structural realism while providing explicit localization of distortions.
Experimental results demonstrate the effectiveness of the proposed framework in accurately assessing structural realism in T2I-generated images.

The model training process is divided into two phases, totaling 50 epochs, to optimize performance and prevent overfitting.
In the first phase (initial training, 20 epochs), the pre-trained ViT and text embedding modules are frozen, and only the remaining parts of the model (such as image and text token branches) are trained to focus on learning task-related features while keeping the pre-trained components stable.
In the second phase (fine-tuning, 30 epochs), all modules are unfrozen, and the entire model is jointly optimized to better perform structural distortion detection and structural quality assessment. 
This two-stage training strategy leverages pre-trained knowledge effectively, progressively refines the model and improves performance efficiently.

During the testing phase, the model receives unseen generated images and their corresponding text prompts as input. 
The image is passed through the ViT to extract image tokens, and the text prompt is passed through the text embedding module to generate text tokens.
After these tokens are concatenated, they are sent to the image token branch and the text token branch for processing respectively. 
The image token branch generates a heat map for accurately locating areas of structural distortion in the image (such as the incorrect number of limbs or facial collapse); the text token branch calculates a score that reflects the structural quality of the image. 
Finally, the model outputs a mask image for locating structural distortions and a structural score, which will be compared with the annotations of the EvalMuse Part 2 dataset to evaluate the accuracy and effectiveness of the model in practical applications.

%% file: sec/X_appendix.tex
\appendix

\section{NTIRE 2025 Organizers}
\label{sec_apd:organizer}
\noindent\textit{\textbf{Title: }}\\ NTIRE 2025 challenge on Text to Image Generation Model Quality Assessment\\
\noindent\textit{\textbf{Members:}}\\ 
Shuhao Han$^{1,2}$ \textit{ (hansh@mail.nankai.edu.cn)}, Haotian Fan$^2$, Fangyuan Kong$^2$, Wenjie Liao$^{1,2}$, Chunle Guo$^1$, Chongyi Li$^1$, Radu Timofte$^{3}$, Liang Li$^2$, Tao Li$^2$, Junhui Cui$^2$, Yunqiu Wang$^2$, Yang Tai$^2$, Jingwei Sun$^2$\\
\noindent\textit{\textbf{Affiliations: }}\\
$^{1}$ Nankai University, China\\
$^{2}$ ByteDance Inc, China\\
$^{3}$ Computer Vision Lab, University of W\"urzburg, Germany

\section{Teams and Affiliations in Alignment Track}
\label{sec_apd:track1_team}

\subsection*{IH-VQA}
\noindent\textit{\textbf{Title:}}\\
Instruction-augmented Multimodal Alignment for Text-Image and Element Matching\\
\noindent\textit{\textbf{Members: }}\\
Jianhui Sun$^1$ \textit{(nimosun@tencent.com)},  Xinli Yue$^2$, Tianyi Wang$^1$, Huan Hou$^1$, Junda Lu$^1$, Xinyang Huang$^1$, Zitang Zhou$^1$\\
\noindent\textit{\textbf{Affiliations: }}\\
$^1$ WeChat\\
$^2$ Wuhan University\\

\subsection*{Evalthon}
\noindent\textit{\textbf{Title:}}\\
Enhancing the Capabilities of Large Vision Lan-guage Models Through the Integration of Statistical Features\\
\noindent\textit{\textbf{Members: }}\\
Zijian Zhang$^1$ \textit{(zhangzijian14@meituan.com)},  Xuhui Zheng$^1$,Xuecheng Wu$^1$, Chong Peng$^1$, Xuezhi Cao$^1$\\
\noindent\textit{\textbf{Affiliations: }}\\
$^1$ Meituan\\

\subsection*{HCMUS}
\noindent\textit{\textbf{Title:}}\\
Fine-Tuning Qwen-VL Family Models with LoRA for Text-to-Image Qual-ity Assessment\\
\noindent\textit{\textbf{Members: }}\\
Trong-Hieu Nguyen-Mau$^1$ \textit{(nmthieu@selab.hcmus.edu.vn)}, Minh-Hoang Le$^1$, Minh-Khoa Le-Phan$^1$, Duy-Nam Ly$^1$, Hai-Dang Nguyen$^1$, Minh-Triet Tran$^1$\\
\noindent\textit{\textbf{Affiliations: }}\\
$^1$ University of Science, VNU-HCM\\

\subsection*{MICV}
\noindent\textit{\textbf{Title:}}\\
A Prior Information-Guided Multi-modal Approach\\
\noindent\textit{\textbf{Members: }}\\
Jun Lan$^1$ 
\textit{(lanjun\_yelan@163.com)}, Yukang Lin$^1$, Yan Hong$^1$, Chuanbiao Song$^1$, Siyuan Li$^1$ \\
\noindent\textit{\textbf{Affiliations: }}\\
$^1$ MICV\\

\subsection*{SJTU-MMLab}
\noindent\textit{\textbf{Title:}}\\
Evaluation of Image-Text Alignment through Multimodal Large LanguageModel\\
\noindent\textit{\textbf{Members: }}\\
Zhichao Zhang$^1$ \textit{(liquortect@sjtu.edu.cn)}, Xinyue Li$^1$, Wei Sun$^1$, Zicheng Zhang$^1$, Yunhao Li$^1$,Xiaohong Liu$^1$, Guangtao Zhai$^1$ \\
\noindent\textit{\textbf{Affiliations: }}\\
$^1$ Shanghai Jiao Tong University\\

\subsection*{SJTUMM}
\noindent\textit{\textbf{Title:}}\\
Multi-Encoder Framework with Quality Assessment via Cross-Modal Alignment and LLM Adaptation\\
\noindent\textit{\textbf{Members: }}\\
Zitong Xu$^1$ \textit{(2766998042@qq.com)}, HuiyuDuan$^1$, JiaruiWang$^1$, GuangjiMa$^1$, LiuYang$^1$, LuLiu$^1$, QiangHu$^1$, Xiongkuo Min$^1$, Guangtao Zhai$^1$ \\
\noindent\textit{\textbf{Affiliations: }}\\
$^1$ Shanghai Jiao Tong University\\

\subsection*{WT}
\noindent\textit{\textbf{Title:}}\\
Using MLLM to Evaluate Image-Text Alignment through SFT and Score Distribution\\
\noindent\textit{\textbf{Members: }}\\
Zichuan Wang$^1$ \textit{(wangzichuan2024@ia.ac.cn)}, Zhenchen Tang$^1$, Bo Peng$^1$, Jing Dong$^1$ \\
\noindent\textit{\textbf{Affiliations: }}\\
$^1$ Institute of Automation, Chinese Academy of Sciences\\

\subsection*{YAG}
\noindent\textit{\textbf{Title:}}\\
A Multi-Level Granularity AIGC Quality Assessment Method Based on Multi-modal Understanding\\
\noindent\textit{\textbf{Members: }}\\
Fengbin Guan$^1$ \textit{(guanfb@mail.ustc.edu.cn)}, Zihao Yu$^1$, Yiting Lu$^1$, Wei Luo$^1$, Xin Li$^1$\\
\noindent\textit{\textbf{Affiliations: }}\\
$^1$ University of Science and Technology of China\\

\subsection*{SPRank}
\noindent\textit{\textbf{Title:}}\\
Qwen-Assisted Image-Text Alignment Scoring\\
\noindent\textit{\textbf{Members: }}\\
Minhao Lin$^1$ \textit{(23080232@hdu.edu.cn)}, Haofeng Chen$^1$\\
\noindent\textit{\textbf{Affiliations: }}\\
$^1$ Hangzhou Dianzi University,\\

\subsection*{AIIG}
\noindent\textit{\textbf{Title:}}\\
Fine-Grained Image-Text Alignment Evaluation\\
\noindent\textit{\textbf{Members: }}\\
Xuanxuan He$^1$ \textit{(2416273879@qq.com)}, Kele Xu$^1$, Qisheng Xu$^1$, Zijian Gao$^1$, Tianjiao Wan$^1$ \\
\noindent\textit{\textbf{Affiliations: }}\\
$^1$ National University of Defense Technology\\

\subsection*{Joe1007}
\noindent\textit{\textbf{Title:}}\\
Enhanced FGA-BLIP2 Model with Contrastive Learning\\
\noindent\textit{\textbf{Members: }}\\
Bo-Cheng, Qiu$^1$ \textit{(e6134048@gs.ncku.edu.tw)}, Chih-Chung Hsu$^1$, Chia-ming Lee$^1$, Yu-Fan Lin$^1$\\
\noindent\textit{\textbf{Affiliations: }}\\
$^1$ National Cheng Kung University\\

\subsection*{iCOST}
\noindent\textit{\textbf{Title:}}\\
VLM-R1 utilization on the task of matching score evaluation\\
\noindent\textit{\textbf{Members: }}\\
Bo Yu$^1$ \textit{(a7858833@bupt.edu.cn)}, Zehao Wang$^1$, Da Mu$^1$, Mingxiu Chen$^1$, Junkang Fang$^1$, Huamei Sun$^1$, Wending Zhao$^1$ \\
\noindent\textit{\textbf{Affiliations: }}\\
$^1$ Beijing University of Posts and Telecommunications\\

\section{Teams and Affiliations in Structure Track}
\label{sec_apd:track2_team}
\subsection*{HNU-VPAI}
\noindent\textit{\textbf{Title:}}\\
Learning Local Distortion Features for Quality Assessment to Text-to-Image Generation Models\\
\noindent\textit{\textbf{Members: }}\\
Zhiyu Wang$^1$ \textit{(wangzhiyu.wzy1@gmail.com)}, Wang Liu$^1$, Weikang Yu$^2$, Puhong Duan$^1$, Bin Sun$^1$, Xudong Kang$^1$, Shutao Li$^1$ \\
\noindent\textit{\textbf{Affiliations: }}\\
$^1$ Hunan University\\
$^2$ Technical University of Munich\\

\subsection*{OPDAI}
\noindent\textit{\textbf{Title:}}\\
Florence2-Based Multi-Task Architecture for Structural Distortion Detection in AIGC Imagery\\
\noindent\textit{\textbf{Members: }}\\
Shuai He$^1$ \textit{(heshuai.sec@gmail.com)}, Lingzhi Fu$^1$ , Heng Cong$^1$ , Rongyu Zhang$^1$, Jiarong He$^1$  \\
\noindent\textit{\textbf{Affiliations: }}\\
$^1$ OPDAI,Netease Game,Guangzhou China\\

\subsection*{MICV}
\noindent\textit{\textbf{Title:}}\\
Study Group Learning for Structure DistortionDetection in T2I Generation Models\\
\noindent\textit{\textbf{Members: }}\\
Jun Lan$^1$ \textit{(lanjun\_yelan@163.com)}, YukangLin$^1$, YanHong$^1$, ChuanbiaoSong$^1$, Siyuan Li$^1$ \\
\noindent\textit{\textbf{Affiliations: }}\\
$^1$ MICV\\

\subsection*{out of memory}
\noindent\textit{\textbf{Title:}}\\
out of memory\\
\noindent\textit{\textbf{Members: }}\\
Zhishan Qiao$^1$ \textit{(19986244365@163.com)}, Yongqing Huang$^1$ \\
\noindent\textit{\textbf{Affiliations: }}\\
$^1$ -\\

\subsection*{I$^2$ Group}
\noindent\textit{\textbf{Title:}}\\
AIGC Image Quality Assessment and Structure Distortion Detection via Image-Prompt\\
\noindent\textit{\textbf{Members: }}\\
Zewen Chen$^1$ \textit{(chenzewen2022@ia.ac.cn)}, Zhe Pang$^{1,2}$, Juan Wang$^1$, Jian Guo$^3$, Zhizhuo Shao$^2$, Ziyu Feng$^4$, Bing Li$^{1,5}$, Weiming Hu$^{1,2,6}$, Hesong Li$^7$, Dehua Liu$^7$ \\
\noindent\textit{\textbf{Affiliations: }}\\
$^1$State Key Laboratory of Multimodal Artificial Intelligence Systems,Institute of Automation, Chinese Academy of Sciences \\
$^2$School of Artificial Intelligence, University of Chinese Academy of Sciences \\
$^3$Beijing Union University \\
$^4$Beijing Jiaotong University \\
$^5$PeopleAI Inc.Beijing, China \\
$^6$School of Information Science and Technology, ShanghaiTech University \\
$^7$Transsion Inc.

\subsection*{Brute Force Wins}
\noindent\textit{\textbf{Title:}}\\
Structure Distortion Detection and Scoring Model Based on SAM and MLLM\\
\noindent\textit{\textbf{Members: }}\\
Zeming Liu$^1$ \textit{(Buuugmaker@163.com)}, Qingsong Xie$^1$, Ruichen Wang$^1$, Zhihao Li$^1$ \\
\noindent\textit{\textbf{Affiliations: }}\\
$^1$  OPPO AI Center\\

\subsection*{WecanEvalAIGC}
\noindent\textit{\textbf{Title:}}\\
VLM-R1 utilization on the task of matching score evaluation\\
\noindent\textit{\textbf{Members: }}\\
Jun Luo$^1$ \textit{(wuguo.lj@antgroup.com)}, Yuqi Liang$^1$, Jianqi Bi$^1$ \\
\noindent\textit{\textbf{Affiliations: }}\\
$^1$ Antgroup\\

\subsection*{Tenryu Babu}
\noindent\textit{\textbf{Title:}}\\
Attention-Driven Structure Quality Assessment for Text-to-Image Gener-ation\\
\noindent\textit{\textbf{Members: }}\\
Junfeng Yang$^1$ \textit{(b12100031@hnu.edu.cn)}, Can Li$^1$ , Jing Fu$^1$ , Hongwei Xu$^1$ , Mingrui Long$^1$, Lulin Tang  \\
\noindent\textit{\textbf{Affiliations: }}\\
$^1$ Xiangjiang Laboratory, Hunan University of Technology and Business\\

%% file: main.bbl
\begin{thebibliography}{78}
\providecommand{\natexlab}[1]{#1}
\providecommand{\url}[1]{\texttt{#1}}
\expandafter\ifx\csname urlstyle\endcsname\relax
  \providecommand{\doi}[1]{doi: #1}\else
  \providecommand{\doi}{doi: \begingroup \urlstyle{rm}\Url}\fi

\bibitem[Bai et~al.(2025{\natexlab{a}})Bai, Chen, Liu, Wang, Ge, Song, Dang, Wang, Wang, Tang, et~al.]{bai2025qwen2}
Shuai Bai, Keqin Chen, Xuejing Liu, Jialin Wang, Wenbin Ge, Sibo Song, Kai Dang, Peng Wang, Shijie Wang, Jun Tang, et~al.
\newblock Qwen2. 5-vl technical report.
\newblock \emph{arXiv preprint arXiv:2502.13923}, 2025{\natexlab{a}}.

\bibitem[Bai et~al.(2025{\natexlab{b}})Bai, Chen, Liu, Wang, Ge, Song, Dang, Wang, Wang, Tang, et~al.]{qwen2.5vl}
Shuai Bai, Keqin Chen, Xuejing Liu, Jialin Wang, Wenbin Ge, Sibo Song, Kai Dang, Peng Wang, Shijie Wang, Jun Tang, et~al.
\newblock Qwen2. 5-vl technical report.
\newblock \emph{arXiv preprint arXiv:2502.13923}, 2025{\natexlab{b}}.

\bibitem[Chen and Guestrin(2016)]{chen2016xgboost}
Tianqi Chen and Carlos Guestrin.
\newblock Xgboost: A scalable tree boosting system.
\newblock In \emph{Proceedings of the 22nd acm sigkdd international conference on knowledge discovery and data mining}, pages 785--794, 2016.

\bibitem[Chen et~al.(2022{\natexlab{a}})Chen, Liu, Zhang, Ye, Yang, and Wu]{chen2022altclip}
Zhongzhi Chen, Guang Liu, Bo-Wen Zhang, Fulong Ye, Qinghong Yang, and Ledell Wu.
\newblock Altclip: Altering the language encoder in clip for extended language capabilities.
\newblock \emph{arXiv preprint arXiv:2211.06679}, 2022{\natexlab{a}}.

\bibitem[Chen et~al.(2022{\natexlab{b}})Chen, Wang, Li, Yuan, Xiong, Cheng, and Hu]{chen2022teacher}
Zewen Chen, Juan Wang, Bing Li, Chunfeng Yuan, Weihua Xiong, Rui Cheng, and Weiming Hu.
\newblock Teacher-guided learning for blind image quality assessment.
\newblock In \emph{Proceedings of the Asian Conference on Computer Vision}, pages 2457--2474, 2022{\natexlab{b}}.

\bibitem[Chen et~al.(2024{\natexlab{a}})Chen, Wang, Wang, Xu, Xiong, Zeng, Guo, Wang, Yuan, Li, et~al.]{chen2024seagull}
Zewen Chen, Juan Wang, Wen Wang, Sunhan Xu, Hang Xiong, Yun Zeng, Jian Guo, Shuxun Wang, Chunfeng Yuan, Bing Li, et~al.
\newblock Seagull: No-reference image quality assessment for regions of interest via vision-language instruction tuning.
\newblock \emph{arXiv preprint arXiv:2411.10161}, 2024{\natexlab{a}}.

\bibitem[Chen et~al.(2024{\natexlab{b}})Chen, Wang, Cao, Liu, Gao, Cui, Zhu, Ye, Tian, Liu, et~al.]{internvl2.5}
Zhe Chen, Weiyun Wang, Yue Cao, Yangzhou Liu, Zhangwei Gao, Erfei Cui, Jinguo Zhu, Shenglong Ye, Hao Tian, Zhaoyang Liu, et~al.
\newblock Expanding performance boundaries of open-source multimodal models with model, data, and test-time scaling.
\newblock \emph{arXiv preprint arXiv:2412.05271}, 2024{\natexlab{b}}.

\bibitem[Chen et~al.(2025{\natexlab{a}})Chen, Liu, Gong, Wang, Sun, Wu, Timofte, Zhang, et~al.]{ntire2025srx4}
Zheng Chen, Kai Liu, Jue Gong, Jingkai Wang, Lei Sun, Zongwei Wu, Radu Timofte, Yulun Zhang, et~al.
\newblock {NTIRE} 2025 challenge on image super-resolution (×4): Methods and results.
\newblock In \emph{Proceedings of the IEEE/CVF Conference on Computer Vision and Pattern Recognition (CVPR) Workshops}, 2025{\natexlab{a}}.

\bibitem[Chen et~al.(2025{\natexlab{b}})Chen, Qin, Wang, Yuan, Li, Hu, and Wang]{chen2024promptiqa}
Zewen Chen, Haina Qin, Juan Wang, Chunfeng Yuan, Bing Li, Weiming Hu, and Liang Wang.
\newblock Promptiqa: Boosting the performance and generalization for no-reference image quality assessment via prompts.
\newblock In \emph{European Conference on Computer Vision}, pages 247--264. Springer, 2025{\natexlab{b}}.

\bibitem[Chen et~al.(2025{\natexlab{c}})Chen, Wang, Liu, Gong, Sun, Wu, Timofte, Zhang, et~al.]{ntire2025face}
Zheng Chen, Jingkai Wang, Kai Liu, Jue Gong, Lei Sun, Zongwei Wu, Radu Timofte, Yulun Zhang, et~al.
\newblock {NTIRE} 2025 challenge on real-world face restoration: Methods and results.
\newblock In \emph{Proceedings of the IEEE/CVF Conference on Computer Vision and Pattern Recognition (CVPR) Workshops}, 2025{\natexlab{c}}.

\bibitem[Conde et~al.(2025{\natexlab{a}})Conde, Timofte, et~al.]{ntire2025raw}
Marcos Conde, Radu Timofte, et~al.
\newblock {NTIRE} 2025 challenge on raw image restoration and super-resolution.
\newblock In \emph{Proceedings of the IEEE/CVF Conference on Computer Vision and Pattern Recognition (CVPR) Workshops}, 2025{\natexlab{a}}.

\bibitem[Conde et~al.(2025{\natexlab{b}})Conde, Timofte, et~al.]{ntire2025rawrgb}
Marcos Conde, Radu Timofte, et~al.
\newblock Raw image reconstruction from {RGB} on smartphones. {NTIRE} 2025 challenge report.
\newblock In \emph{Proceedings of the IEEE/CVF Conference on Computer Vision and Pattern Recognition (CVPR) Workshops}, 2025{\natexlab{b}}.

\bibitem[Ershov et~al.(2025)Ershov, Korchagin, Khalin, Panshin, Terekhin, Zaychenkova, Lobarev, Plokhotnyuk, Abramov, Zhdanov, Dorogova, Mamedov, Banic, Perevozchikov, Timofte, et~al.]{ntire2025night}
Egor Ershov, Sergey Korchagin, Alexei Khalin, Artyom Panshin, Arseniy Terekhin, Ekaterina Zaychenkova, Georgiy Lobarev, Vsevolod Plokhotnyuk, Denis Abramov, Elisey Zhdanov, Sofia Dorogova, Yasin Mamedov, Nikola Banic, Georgii Perevozchikov, Radu Timofte, et~al.
\newblock {NTIRE} 2025 challenge on night photography rendering.
\newblock In \emph{Proceedings of the IEEE/CVF Conference on Computer Vision and Pattern Recognition (CVPR) Workshops}, 2025.

\bibitem[Fang et~al.(2022)Fang, Wang, Xie, Sun, Wu, Wang, Huang, Wang, and Cao]{EVA}
Yuxin Fang, Wen Wang, Binhui Xie, Quan Sun, Ledell Wu, Xinggang Wang, Tiejun Huang, Xinlong Wang, and Yue Cao.
\newblock Eva: Exploring the limits of masked visual representation learning at scale.
\newblock \emph{arXiv preprint arXiv:2211.07636}, 2022.

\bibitem[Fu et~al.(2025)Fu, Qiu, Fu, Timofte, Sebe, Yang, Van~Gool, et~al.]{ntire2025cross}
Yuqian Fu, Xingyu Qiu, Bin Ren~Yanwei Fu, Radu Timofte, Nicu Sebe, Ming-Hsuan Yang, Luc Van~Gool, et~al.
\newblock {NTIRE} 2025 challenge on cross-domain few-shot object detection: Methods and results.
\newblock In \emph{Proceedings of the IEEE/CVF Conference on Computer Vision and Pattern Recognition (CVPR) Workshops}, 2025.

\bibitem[Guo et~al.(2025)Guo, Yang, Zhang, Song, Zhang, Xu, Zhu, Ma, Wang, Bi, et~al.]{guo2025deepseek}
Daya Guo, Dejian Yang, Haowei Zhang, Junxiao Song, Ruoyu Zhang, Runxin Xu, Qihao Zhu, Shirong Ma, Peiyi Wang, Xiao Bi, et~al.
\newblock Deepseek-r1: Incentivizing reasoning capability in llms via reinforcement learning.
\newblock \emph{arXiv preprint arXiv:2501.12948}, 2025.

\bibitem[Han et~al.(2024)Han, Fan, Fu, Li, Li, Cui, Wang, Tai, Sun, Guo, and Li]{han2024evalmuse}
Shuhao Han, Haotian Fan, Jiachen Fu, Liang Li, Tao Li, Junhui Cui, Yunqiu Wang, Yang Tai, Jingwei Sun, Chunle Guo, and Chongyi Li.
\newblock Evalmuse-40k: A reliable and fine-grained benchmark with comprehensive human annotations for text-to-image generation model evaluation, 2024.

\bibitem[Han et~al.(2025)Han, Fan, Kong, Liao, Guo, Li, Timofte, et~al.]{ntire2025text}
Shuhao Han, Haotian Fan, Fangyuan Kong, Wenjie Liao, Chunle Guo, Chongyi Li, Radu Timofte, et~al.
\newblock {NTIRE} 2025 challenge on text to image generation model quality assessment.
\newblock In \emph{Proceedings of the IEEE/CVF Conference on Computer Vision and Pattern Recognition (CVPR) Workshops}, 2025.

\bibitem[He et~al.(2022)He, Yang, Zhang, Shan, and Chen]{he2022latent}
Yingqing He, Tianyu Yang, Yong Zhang, Ying Shan, and Qifeng Chen.
\newblock Latent video diffusion models for high-fidelity video generation with arbitrary lengths.
\newblock \emph{arXiv preprint arXiv:2211.13221}, 2022.

\bibitem[Hu et~al.(2022)Hu, Shen, Wallis, Allen-Zhu, Li, Wang, Wang, and Chen]{hu2022lora}
Edward~J Hu, Yelong Shen, Phillip Wallis, Zeyuan Allen-Zhu, Yuanzhi Li, Shean Wang, Lu Wang, and Weizhu Chen.
\newblock Lo{RA}: Low-rank adaptation of large language models.
\newblock In \emph{ICLR}, 2022.

\bibitem[Hu et~al.(2023)Hu, Liu, Kasai, Wang, Ostendorf, Krishna, and Smith]{hu2023tifa}
Yushi Hu, Benlin Liu, Jungo Kasai, Yizhong Wang, Mari Ostendorf, Ranjay Krishna, and Noah~A Smith.
\newblock Tifa: Accurate and interpretable text-to-image faithfulness evaluation with question answering.
\newblock In \emph{Proceedings of the IEEE/CVF International Conference on Computer Vision}, pages 20406--20417, 2023.

\bibitem[Jain et~al.(2025)Jain, Wu, Zou, Florentin, Turbell, Siddhartha, Timofte, et~al.]{ntire2025vqe}
Varun Jain, Zongwei Wu, Quan Zou, Louis Florentin, Henrik Turbell, Sandeep Siddhartha, Radu Timofte, et~al.
\newblock {NTIRE} 2025 challenge on video quality enhancement for video conferencing: Datasets, methods and results.
\newblock In \emph{Proceedings of the IEEE/CVF Conference on Computer Vision and Pattern Recognition (CVPR) Workshops}, 2025.

\bibitem[Kirstain et~al.(2024)Kirstain, Polyak, Singer, Matiana, Penna, and Levy]{kirstain2024pick}
Yuval Kirstain, Adam Polyak, Uriel Singer, Shahbuland Matiana, Joe Penna, and Omer Levy.
\newblock Pick-a-pic: An open dataset of user preferences for text-to-image generation.
\newblock \emph{Advances in Neural Information Processing Systems}, 36, 2024.

\bibitem[Lee et~al.(2025)Lee, Park, Canelo, Park, Kim, Chun, Jin, Li, Guo, Timofte, et~al.]{ntire2025ebhdr}
Sangmin Lee, Eunpil Park, Angel Canelo, Hyunhee Park, Youngjo Kim, Hyungju Chun, Xin Jin, Chongyi Li, Chun-Le Guo, Radu Timofte, et~al.
\newblock {NTIRE} 2025 challenge on efficient burst hdr and restoration: Datasets, methods, and results.
\newblock In \emph{Proceedings of the IEEE/CVF Conference on Computer Vision and Pattern Recognition (CVPR) Workshops}, 2025.

\bibitem[Li et~al.(2024{\natexlab{a}})Li, Lin, Pathak, Li, Fei, Wu, Ling, Xia, Zhang, Neubig, et~al.]{li2024genai}
Baiqi Li, Zhiqiu Lin, Deepak Pathak, Jiayao Li, Yixin Fei, Kewen Wu, Tiffany Ling, Xide Xia, Pengchuan Zhang, Graham Neubig, et~al.
\newblock Genai-bench: Evaluating and improving compositional text-to-visual generation.
\newblock \emph{arXiv preprint arXiv:2406.13743}, 2024{\natexlab{a}}.

\bibitem[Li et~al.(2024{\natexlab{b}})Li, Zhang, Guo, Zhang, Li, Zhang, Zhang, Zhang, Li, Liu, et~al.]{li2024llava}
Bo Li, Yuanhan Zhang, Dong Guo, Renrui Zhang, Feng Li, Hao Zhang, Kaichen Zhang, Peiyuan Zhang, Yanwei Li, Ziwei Liu, et~al.
\newblock Llava-onevision: Easy visual task transfer.
\newblock \emph{arXiv preprint arXiv:2408.03326}, 2024{\natexlab{b}}.

\bibitem[Li et~al.(2023{\natexlab{a}})Li, Zhang, Wu, Sun, Min, Liu, Zhai, and Lin]{li2023agiqa}
Chunyi Li, Zicheng Zhang, Haoning Wu, Wei Sun, Xiongkuo Min, Xiaohong Liu, Guangtao Zhai, and Weisi Lin.
\newblock Agiqa-3k: An open database for ai-generated image quality assessment.
\newblock \emph{IEEE Transactions on Circuits and Systems for Video Technology}, 2023{\natexlab{a}}.

\bibitem[Li et~al.(2024{\natexlab{c}})Li, Kou, Gao, Cao, Sun, Zhang, Zhou, Zhang, Wu, Zhang, Liu, Min, and Zhai]{AIGIQA-20K}
Chunyi Li, Tengchuan Kou, Yixuan Gao, Yuqin Cao, Wei Sun, Zicheng Zhang, Yingjie Zhou, Zhichao Zhang, Haoning Wu, Weixia Zhang, Xiaohong Liu, Xiongkuo Min, and Guangtao Zhai.
\newblock Aigiqa-20k: A large database for ai-generated image quality assessment.
\newblock In \emph{Proceedings of the IEEE/CVF Conference on Computer Vision and Pattern Recognition Workshops}, 2024{\natexlab{c}}.

\bibitem[Li et~al.(2022)Li, Li, Xiong, and Hoi]{li2022blip}
Junnan Li, Dongxu Li, Caiming Xiong, and Steven Hoi.
\newblock Blip: Bootstrapping language-image pre-training for unified vision-language understanding and generation.
\newblock In \emph{International conference on machine learning}, pages 12888--12900. PMLR, 2022.

\bibitem[Li et~al.(2023{\natexlab{b}})Li, Li, Savarese, and Hoi]{li2023blip2}
Junnan Li, Dongxu Li, Silvio Savarese, and Steven Hoi.
\newblock Blip-2: Bootstrapping language-image pre-training with frozen image encoders and large language models.
\newblock In \emph{International conference on machine learning}, pages 19730--19742. PMLR, 2023{\natexlab{b}}.

\bibitem[Li et~al.(2025{\natexlab{a}})Li, Jin, Jin, Wu, Li, Wang, Yang, Li, Chen, Wen, Tan, Timofte, et~al.]{ntire2025day}
Xin Li, Yeying Jin, Xin Jin, Zongwei Wu, Bingchen Li, Yufei Wang, Wenhan Yang, Yu Li, Zhibo Chen, Bihan Wen, Robby Tan, Radu Timofte, et~al.
\newblock {NTIRE} 2025 challenge on day and night raindrop removal for dual-focused images: Methods and results.
\newblock In \emph{Proceedings of the IEEE/CVF Conference on Computer Vision and Pattern Recognition (CVPR) Workshops}, 2025{\natexlab{a}}.

\bibitem[Li et~al.(2025{\natexlab{b}})Li, Wang, Li, Yuan, Shao, Yao, Sun, Zhou, Timofte, and Chen]{ntire2025shortugc_data}
Xin Li, Xijun Wang, Bingchen Li, Kun Yuan, Yizhen Shao, Suhang Yao, Ming Sun, Chao Zhou, Radu Timofte, and Zhibo Chen.
\newblock {NTIRE} 2025 challenge on short-form ugc video quality assessment and enhancement: Kwaisr dataset and study.
\newblock In \emph{Proceedings of the IEEE/CVF Conference on Computer Vision and Pattern Recognition (CVPR) Workshops}, 2025{\natexlab{b}}.

\bibitem[Li et~al.(2025{\natexlab{c}})Li, Yuan, Li, Guan, Shao, Yu, Wang, Lu, Luo, Yao, Sun, Zhou, Chen, Timofte, et~al.]{ntire2025shortugc}
Xin Li, Kun Yuan, Bingchen Li, Fengbin Guan, Yizhen Shao, Zihao Yu, Xijun Wang, Yiting Lu, Wei Luo, Suhang Yao, Ming Sun, Chao Zhou, Zhibo Chen, Radu Timofte, et~al.
\newblock {NTIRE} 2025 challenge on short-form ugc video quality assessment and enhancement: Methods and results.
\newblock In \emph{Proceedings of the IEEE/CVF Conference on Computer Vision and Pattern Recognition (CVPR) Workshops}, 2025{\natexlab{c}}.

\bibitem[Liang et~al.(2025)Liang, Timofte, Yi, Zhang, Liu, Sun, Wu, Zhang, Zeng, Zhang, et~al.]{ntire2025raim}
Jie Liang, Radu Timofte, Qiaosi Yi, Zhengqiang Zhang, Shuaizheng Liu, Lingchen Sun, Rongyuan Wu, Xindong Zhang, Hui Zeng, Lei Zhang, et~al.
\newblock {NTIRE} 2025 the 2nd restore any image model {(RAIM)} in the wild challenge.
\newblock In \emph{Proceedings of the IEEE/CVF Conference on Computer Vision and Pattern Recognition (CVPR) Workshops}, 2025.

\bibitem[Liang et~al.(2024)Liang, He, Li, Li, Klimovskiy, Carolan, Sun, Pont-Tuset, Young, Yang, et~al.]{liang2024rich}
Youwei Liang, Junfeng He, Gang Li, Peizhao Li, Arseniy Klimovskiy, Nicholas Carolan, Jiao Sun, Jordi Pont-Tuset, Sarah Young, Feng Yang, et~al.
\newblock Rich human feedback for text-to-image generation.
\newblock In \emph{Proceedings of the IEEE/CVF Conference on Computer Vision and Pattern Recognition}, pages 19401--19411, 2024.

\bibitem[Lin et~al.(2024)Lin, Pathak, Li, Li, Xia, Neubig, Zhang, and Ramanan]{lin2024evaluating}
Zhiqiu Lin, Deepak Pathak, Baiqi Li, Jiayao Li, Xide Xia, Graham Neubig, Pengchuan Zhang, and Deva Ramanan.
\newblock Evaluating text-to-visual generation with image-to-text generation.
\newblock In \emph{European Conference on Computer Vision}, pages 366--384. Springer, 2024.

\bibitem[Liu et~al.(2025{\natexlab{a}})Liu, Min, Hu, Zhang, Guo, et~al.]{ntire2025xgc}
Xiaohong Liu, Xiongkuo Min, Qiang Hu, Xiaoyun Zhang, Jie Guo, et~al.
\newblock {NTIRE} 2025 {XGC} quality assessment challenge: Methods and results.
\newblock In \emph{Proceedings of the IEEE/CVF Conference on Computer Vision and Pattern Recognition (CVPR) Workshops}, 2025{\natexlab{a}}.

\bibitem[Liu et~al.(2025{\natexlab{b}})Liu, Wu, Vasluianu, Yan, Ren, Zhang, Gu, Zhang, Zhu, Timofte, et~al.]{ntire2025lowlight}
Xiaoning Liu, Zongwei Wu, Florin-Alexandru Vasluianu, Hailong Yan, Bin Ren, Yulun Zhang, Shuhang Gu, Le Zhang, Ce Zhu, Radu Timofte, et~al.
\newblock {NTIRE} 2025 challenge on low light image enhancement: Methods and results.
\newblock In \emph{Proceedings of the IEEE/CVF Conference on Computer Vision and Pattern Recognition (CVPR) Workshops}, 2025{\natexlab{b}}.

\bibitem[Liu et~al.(2021)Liu, Lin, Cao, Hu, Wei, Zhang, Lin, and Guo]{liu2021swin}
Ze Liu, Yutong Lin, Yue Cao, Han Hu, Yixuan Wei, Zheng Zhang, Stephen Lin, and Baining Guo.
\newblock Swin transformer: Hierarchical vision transformer using shifted windows.
\newblock In \emph{Proceedings of the IEEE/CVF International Conference on Computer Vision}, pages 10012--10022, 2021.

\bibitem[Lu et~al.(2024)Lu, Li, Chen, Xu, Luo, Zhang, and Ye]{ovis}
Shiyin Lu, Yang Li, Qing-Guo Chen, Zhao Xu, Weihua Luo, Kaifu Zhang, and Han-Jia Ye.
\newblock Ovis: Structural embedding alignment for multimodal large language model.
\newblock \emph{arXiv:2405.20797}, 2024.

\bibitem[Radford et~al.(2021)Radford, Kim, Hallacy, Ramesh, Goh, Agarwal, Sastry, Askell, Mishkin, Clark, et~al.]{radford2021clip}
Alec Radford, Jong~Wook Kim, Chris Hallacy, Aditya Ramesh, Gabriel Goh, Sandhini Agarwal, Girish Sastry, Amanda Askell, Pamela Mishkin, Jack Clark, et~al.
\newblock Learning transferable visual models from natural language supervision.
\newblock In \emph{International conference on machine learning}, pages 8748--8763. PMLR, 2021.

\bibitem[Ren et~al.(2025)Ren, Guo, Sun, Wu, Timofte, Li, et~al.]{ntire2025esr}
Bin Ren, Hang Guo, Lei Sun, Zongwei Wu, Radu Timofte, Yawei Li, et~al.
\newblock The tenth {NTIRE} 2025 efficient super-resolution challenge report.
\newblock In \emph{Proceedings of the IEEE/CVF Conference on Computer Vision and Pattern Recognition (CVPR) Workshops}, 2025.

\bibitem[Ronneberger et~al.(2015)Ronneberger, Fischer, and Brox]{ronneberger2015u}
Olaf Ronneberger, Philipp Fischer, and Thomas Brox.
\newblock U-net: Convolutional networks for biomedical image segmentation.
\newblock In \emph{Medical image computing and computer-assisted intervention--MICCAI 2015: 18th international conference, Munich, Germany, October 5-9, 2015, proceedings, part III 18}, pages 234--241. Springer, 2015.

\bibitem[Safonov et~al.(2025)Safonov, Bryntsev, Moskalenko, Kulikov, Vatolin, Timofte, et~al.]{ntire2025ugc}
Nickolay Safonov, Alexey Bryntsev, Andrey Moskalenko, Dmitry Kulikov, Dmitriy Vatolin, Radu Timofte, et~al.
\newblock {NTIRE} 2025 challenge on {UGC} video enhancement: Methods and results.
\newblock In \emph{Proceedings of the IEEE/CVF Conference on Computer Vision and Pattern Recognition (CVPR) Workshops}, 2025.

\bibitem[Sun et~al.(2025{\natexlab{a}})Sun, Alfarano, Duan, Su, Wang, Shi, Timofte, Paudel, Van~Gool, et~al.]{ntire2025event}
Lei Sun, Andrea Alfarano, Peiqi Duan, Shaolin Su, Kaiwei Wang, Boxin Shi, Radu Timofte, Danda~Pani Paudel, Luc Van~Gool, et~al.
\newblock {NTIRE} 2025 challenge on event-based image deblurring: Methods and results.
\newblock In \emph{Proceedings of the IEEE/CVF Conference on Computer Vision and Pattern Recognition (CVPR) Workshops}, 2025{\natexlab{a}}.

\bibitem[Sun et~al.(2025{\natexlab{b}})Sun, Guo, Ren, Van~Gool, Timofte, Li, et~al.]{ntire2025denoising}
Lei Sun, Hang Guo, Bin Ren, Luc Van~Gool, Radu Timofte, Yawei Li, et~al.
\newblock The tenth ntire 2025 image denoising challenge report.
\newblock In \emph{Proceedings of the IEEE/CVF Conference on Computer Vision and Pattern Recognition (CVPR) Workshops}, 2025{\natexlab{b}}.

\bibitem[Sun et~al.(2023)Sun, Min, Tu, Ma, and Zhai]{sun2023blind}
Wei Sun, Xiongkuo Min, Danyang Tu, Siwei Ma, and Guangtao Zhai.
\newblock Blind quality assessment for in-the-wild images via hierarchical feature fusion and iterative mixed database training.
\newblock \emph{IEEE Journal of Selected Topics in Signal Processing}, 2023.

\bibitem[Tschannen et~al.(2025)Tschannen, Gritsenko, Wang, Naeem, Alabdulmohsin, Parthasarathy, Evans, Beyer, Xia, Mustafa, et~al.]{tschannen2025siglip}
Michael Tschannen, Alexey Gritsenko, Xiao Wang, Muhammad~Ferjad Naeem, Ibrahim Alabdulmohsin, Nikhil Parthasarathy, Talfan Evans, Lucas Beyer, Ye Xia, Basil Mustafa, et~al.
\newblock Siglip 2: Multilingual vision-language encoders with improved semantic understanding, localization, and dense features.
\newblock \emph{arXiv preprint arXiv:2502.14786}, 2025.

\bibitem[Vasluianu et~al.(2025{\natexlab{a}})Vasluianu, Seizinger, Zhou, Chen, Wu, Timofte, et~al.]{ntire2025shadow}
Florin-Alexandru Vasluianu, Tim Seizinger, Zhuyun Zhou, Cailian Chen, Zongwei Wu, Radu Timofte, et~al.
\newblock {NTIRE} 2025 image shadow removal challenge report.
\newblock In \emph{Proceedings of the IEEE/CVF Conference on Computer Vision and Pattern Recognition (CVPR) Workshops}, 2025{\natexlab{a}}.

\bibitem[Vasluianu et~al.(2025{\natexlab{b}})Vasluianu, Seizinger, Zhou, Wu, Timofte, et~al.]{ntire2025ambient}
Florin-Alexandru Vasluianu, Tim Seizinger, Zhuyun Zhou, Zongwei Wu, Radu Timofte, et~al.
\newblock {NTIRE} 2025 ambient lighting normalization challenge.
\newblock In \emph{Proceedings of the IEEE/CVF Conference on Computer Vision and Pattern Recognition (CVPR) Workshops}, 2025{\natexlab{b}}.

\bibitem[Wang et~al.(2023{\natexlab{a}})Wang, Chen, Yuan, Li, Ma, and Hu]{wang2023Hierarchical}
Juan Wang, Zewen Chen, Chunfeng Yuan, Bing Li, Wentao Ma, and Weiming Hu.
\newblock Hierarchical curriculum learning for no-reference image quality assessment.
\newblock \emph{International Journal of Computer Vision}, pages 1--20, 2023{\natexlab{a}}.

\bibitem[Wang et~al.(2023{\natexlab{b}})Wang, Duan, Liu, Chen, Min, and Zhai]{wang2023aigciqa2023}
Jiarui Wang, Huiyu Duan, Jing Liu, Shi Chen, Xiongkuo Min, and Guangtao Zhai.
\newblock Aigciqa2023: A large-scale image quality assessment database for ai generated images: from the perspectives of quality, authenticity and correspondence.
\newblock In \emph{CAAI International Conference on Artificial Intelligence}, pages 46--57. Springer, 2023{\natexlab{b}}.

\bibitem[Wang et~al.(2025{\natexlab{a}})Wang, Liang, Zhang, Tian, Wang, Li, Yang, Timofte, Guo, et~al.]{ntire2025lightfield}
Yingqian Wang, Zhengyu Liang, Fengyuan Zhang, Lvli Tian, Longguang Wang, Juncheng Li, Jungang Yang, Radu Timofte, Yulan Guo, et~al.
\newblock {NTIRE} 2025 challenge on light field image super-resolution: Methods and results.
\newblock In \emph{Proceedings of the IEEE/CVF Conference on Computer Vision and Pattern Recognition (CVPR) Workshops}, 2025{\natexlab{a}}.

\bibitem[Wang et~al.(2025{\natexlab{b}})Wang, Zang, Li, Jin, and Wang]{wang2025unified}
Yibin Wang, Yuhang Zang, Hao Li, Cheng Jin, and Jiaqi Wang.
\newblock Unified reward model for multimodal understanding and generation.
\newblock \emph{arXiv preprint arXiv:2503.05236}, 2025{\natexlab{b}}.

\bibitem[Wang et~al.(2022)Wang, Montoya, Munechika, Yang, Hoover, and Chau]{wang2022diffusiondb}
Zijie~J Wang, Evan Montoya, David Munechika, Haoyang Yang, Benjamin Hoover, and Duen~Horng Chau.
\newblock Diffusiondb: A large-scale prompt gallery dataset for text-to-image generative models.
\newblock \emph{arXiv preprint arXiv:2210.14896}, 2022.

\bibitem[Wiles et~al.(2024)Wiles, Zhang, Albuquerque, Kaji{\'c}, Wang, Bugliarello, Onoe, Papalampidi, Ktena, Knutsen, et~al.]{wiles2024revisiting}
Olivia Wiles, Chuhan Zhang, Isabela Albuquerque, Ivana Kaji{\'c}, Su Wang, Emanuele Bugliarello, Yasumasa Onoe, Pinelopi Papalampidi, Ira Ktena, Chris Knutsen, et~al.
\newblock Revisiting text-to-image evaluation with gecko: On metrics, prompts, and human ratings.
\newblock \emph{arXiv preprint arXiv:2404.16820}, 2024.

\bibitem[Wu et~al.(2023{\natexlab{a}})Wu, Zhang, Zhang, Chen, Liao, Li, Gao, Wang, Zhang, Sun, et~al.]{wu2023q}
Haoning Wu, Zicheng Zhang, Weixia Zhang, Chaofeng Chen, Liang Liao, Chunyi Li, Yixuan Gao, Annan Wang, Erli Zhang, Wenxiu Sun, et~al.
\newblock Q-align: Teaching lmms for visual scoring via discrete text-defined levels.
\newblock \emph{arXiv preprint arXiv:2312.17090}, 2023{\natexlab{a}}.

\bibitem[Wu et~al.(2023{\natexlab{b}})Wu, Zhang, Zhang, Chen, Liao, Li, Gao, Wang, Zhang, Sun, et~al.]{wu2023qalign}
Haoning Wu, Zicheng Zhang, Weixia Zhang, Chaofeng Chen, Liang Liao, Chunyi Li, Yixuan Gao, Annan Wang, Erli Zhang, Wenxiu Sun, et~al.
\newblock Q-align: Teaching lmms for visual scoring via discrete text-defined levels.
\newblock \emph{arXiv preprint arXiv:2312.17090}, 2023{\natexlab{b}}.

\bibitem[Wu et~al.(2023{\natexlab{c}})Wu, Sun, Zhu, Zhao, and Li]{wu2023human}
Xiaoshi Wu, Keqiang Sun, Feng Zhu, Rui Zhao, and Hongsheng Li.
\newblock Human preference score: Better aligning text-to-image models with human preference.
\newblock In \emph{Proceedings of the IEEE/CVF International Conference on Computer Vision}, pages 2096--2105, 2023{\natexlab{c}}.

\bibitem[Wu et~al.(2024)Wu, Chen, Pan, Liu, Liu, Dai, Gao, Ma, Wu, Wang, et~al.]{deepseekvl2}
Zhiyu Wu, Xiaokang Chen, Zizheng Pan, Xingchao Liu, Wen Liu, Damai Dai, Huazuo Gao, Yiyang Ma, Chengyue Wu, Bingxuan Wang, et~al.
\newblock Deepseek-vl2: Mixture-of-experts vision-language models for advanced multimodal understanding.
\newblock \emph{arXiv preprint arXiv:2412.10302}, 2024.

\bibitem[Xiao et~al.(2024)Xiao, Wu, Xu, Dai, Hu, Lu, Zeng, Liu, and Yuan]{xiao2024florence}
Bin Xiao, Haiping Wu, Weijian Xu, Xiyang Dai, Houdong Hu, Yumao Lu, Michael Zeng, Ce Liu, and Lu Yuan.
\newblock Florence-2: Advancing a unified representation for a variety of vision tasks.
\newblock In \emph{Proceedings of the IEEE/CVF Conference on Computer Vision and Pattern Recognition}, pages 4818--4829, 2024.

\bibitem[Xiong et~al.(2024)Xiong, Wu, Tan, Li, Tang, Chen, Li, Ma, and Li]{xiong2024sam2}
Xinyu Xiong, Zihuang Wu, Shuangyi Tan, Wenxue Li, Feilong Tang, Ying Chen, Siying Li, Jie Ma, and Guanbin Li.
\newblock Sam2-unet: Segment anything 2 makes strong encoder for natural and medical image segmentation.
\newblock \emph{arXiv preprint arXiv:2408.08870}, 2024.

\bibitem[Xu et~al.(2024{\natexlab{a}})Xu, Liu, Wu, Tong, Li, Ding, Tang, and Dong]{xu2024imagereward}
Jiazheng Xu, Xiao Liu, Yuchen Wu, Yuxuan Tong, Qinkai Li, Ming Ding, Jie Tang, and Yuxiao Dong.
\newblock Imagereward: Learning and evaluating human preferences for text-to-image generation.
\newblock \emph{Advances in Neural Information Processing Systems}, 36, 2024{\natexlab{a}}.

\bibitem[Xu et~al.(2024{\natexlab{b}})Xu, Liao, Xiao, Chen, Wu, Yan, and Lin]{xu2024boosting}
Kangmin Xu, Liang Liao, Jing Xiao, Chaofeng Chen, Haoning Wu, Qiong Yan, and Weisi Lin.
\newblock Boosting image quality assessment through efficient transformer adaptation with local feature enhancement.
\newblock In \emph{Proceedings of the IEEE/CVF Conference on Computer Vision and Pattern Recognition}, pages 2662--2672, 2024{\natexlab{b}}.

\bibitem[Yan et~al.(2024)Yan, Li, Cai, Hao, Jiang, Hu, and Xie]{yan2024sanity}
Shilin Yan, Ouxiang Li, Jiayin Cai, Yanbin Hao, Xiaolong Jiang, Yao Hu, and Weidi Xie.
\newblock A sanity check for ai-generated image detection.
\newblock \emph{arXiv preprint arXiv:2406.19435}, 2024.

\bibitem[Yang et~al.(2025)Yang, Cai, Ouyang, Vasluianu, Timofte, Ding, Sun, Fu, Li, Ho, Meng, et~al.]{ntire2025reflection}
Kangning Yang, Jie Cai, Ling Ouyang, Florin-Alexandru Vasluianu, Radu Timofte, Jiaming Ding, Huiming Sun, Lan Fu, Jinlong Li, Chiu~Man Ho, Zibo Meng, et~al.
\newblock {NTIRE} 2025 challenge on single image reflection removal in the wild: Datasets, methods and results.
\newblock In \emph{Proceedings of the IEEE/CVF Conference on Computer Vision and Pattern Recognition (CVPR) Workshops}, 2025.

\bibitem[Yang et~al.(2022)Yang, Wu, Shi, Lao, Gong, Cao, Wang, and Yang]{yang2022maniqa}
Sidi Yang, Tianhe Wu, Shuwei Shi, Shanshan Lao, Yuan Gong, Mingdeng Cao, Jiahao Wang, and Yujiu Yang.
\newblock Maniqa: Multi-dimension attention network for no-reference image quality assessment.
\newblock In \emph{Proceedings of the IEEE/CVF Conference on Computer Vision and Pattern Recognition}, pages 1191--1200, 2022.

\bibitem[Ye et~al.(2024)Ye, Xu, Ye, Yan, Hu, Liu, Qian, Zhang, and Huang]{ye2024mplug}
Qinghao Ye, Haiyang Xu, Jiabo Ye, Ming Yan, Anwen Hu, Haowei Liu, Qi Qian, Ji Zhang, and Fei Huang.
\newblock mplug-owl2: Revolutionizing multi-modal large language model with modality collaboration.
\newblock In \emph{Proceedings of the ieee/cvf conference on computer vision and pattern recognition}, pages 13040--13051, 2024.

\bibitem[You et~al.(2025{\natexlab{a}})You, Cai, Gu, Xue, and Dong]{deqascore}
Zhiyuan You, Xin Cai, Jinjin Gu, Tianfan Xue, and Chao Dong.
\newblock Teaching large language models to regress accurate image quality scores using score distribution.
\newblock \emph{arXiv preprint arXiv:2501.11561}, 2025{\natexlab{a}}.

\bibitem[You et~al.(2025{\natexlab{b}})You, Cai, Gu, Xue, and Dong]{you2025teaching}
Zhiyuan You, Xin Cai, Jinjin Gu, Tianfan Xue, and Chao Dong.
\newblock Teaching large language models to regress accurate image quality scores using score distribution.
\newblock \emph{arXiv preprint arXiv:2501.11561}, 2025{\natexlab{b}}.

\bibitem[Yue et~al.(2025)Yue, Sun, Lu, Yao, Xia, Wang, Rao, Lyu, and Deng]{yue2025instruction}
Xinli Yue, JianHui Sun, Junda Lu, Liangchao Yao, Fan Xia, Tianyi Wang, Fengyun Rao, Jing Lyu, and Yuetang Deng.
\newblock Instruction-augmented multimodal alignment for image-text and element matching.
\newblock In \emph{Proceedings of the IEEE/CVF Conference on Computer Vision and Pattern Recognition (CVPR) Workshops}, 2025.

\bibitem[Zama~Ramirez et~al.(2025)Zama~Ramirez, Tosi, Di~Stefano, Timofte, Costanzino, Poggi, Salti, Mattoccia, et~al.]{ntire2025hrdepth}
Pierluigi Zama~Ramirez, Fabio Tosi, Luigi Di~Stefano, Radu Timofte, Alex Costanzino, Matteo Poggi, Samuele Salti, Stefano Mattoccia, et~al.
\newblock {NTIRE} 2025 challenge on hr depth from images of specular and transparent surfaces.
\newblock In \emph{Proceedings of the IEEE/CVF Conference on Computer Vision and Pattern Recognition (CVPR) Workshops}, 2025.

\bibitem[Zhang et~al.(2020)Zhang, Ma, Yan, Deng, and Wang]{zhang2020blind}
Weixia Zhang, Kede Ma, Jia Yan, Dexiang Deng, and Zhou Wang.
\newblock Blind image quality assessment using a deep bilinear convolutional neural network.
\newblock \emph{IEEE Transactions on Circuits and Systems for Video Technology}, 30\penalty0 (1):\penalty0 36--47, 2020.

\bibitem[Zhang et~al.(2023{\natexlab{a}})Zhang, Zhai, Wei, Yang, and Ma]{zhang2023liqe}
Weixia Zhang, Guangtao Zhai, Ying Wei, Xiaokang Yang, and Kede Ma.
\newblock Blind image quality assessment via vision-language correspondence: A multitask learning perspective.
\newblock In \emph{IEEE Conference on Computer Vision and Pattern Recognition}, pages 14071--14081, 2023{\natexlab{a}}.

\bibitem[Zhang et~al.(2023{\natexlab{b}})Zhang, Li, Sun, Liu, Min, and Zhai]{zhang2023perceptual}
Zicheng Zhang, Chunyi Li, Wei Sun, Xiaohong Liu, Xiongkuo Min, and Guangtao Zhai.
\newblock A perceptual quality assessment exploration for aigc images.
\newblock In \emph{2023 IEEE International Conference on Multimedia and Expo Workshops (ICMEW)}, pages 440--445. IEEE, 2023{\natexlab{b}}.

\bibitem[Zhang et~al.(2025)Zhang, Zheng, Wu, Peng, and Cao]{zhang2025tokenfocusvqae}
Zijian Zhang, Xuhui Zheng, Xuecheng Wu, Chong Peng, and Xuezhi Cao.
\newblock Tokenfocus-vqa: Enhancing text-to-image alignment with position-aware focus and multi-perspective aggregations on lvlms, 2025.

\bibitem[Zhou et~al.(2021)Zhou, Yu, and Shi]{zhou2021study}
Yuqian Zhou, Hanchao Yu, and Humphrey Shi.
\newblock Study group learning: Improving retinal vessel segmentation trained with noisy labels.
\newblock In \emph{Medical Image Computing and Computer Assisted Intervention--MICCAI 2021: 24th International Conference, Strasbourg, France, September 27--October 1, 2021, Proceedings, Part I 24}, pages 57--67. Springer, 2021.

\bibitem[Zong et~al.(2023)Zong, Song, and Liu]{zong2023detrs}
Zhuofan Zong, Guanglu Song, and Yu Liu.
\newblock Detrs with collaborative hybrid assignments training.
\newblock In \emph{Proceedings of the IEEE/CVF international conference on computer vision}, pages 6748--6758, 2023.

\end{thebibliography}
